\RequirePackage{fix-cm}
\documentclass[twocolumn]{svjour3}          % twocolumn

\smartqed   % flush right qed marks, e.g. at end of proof

\makeatletter
\def\cl@chapter{\@elt {theorem}}
\makeatother
\usepackage[utf8]{inputenc}
\usepackage{graphicx}
\usepackage{color}
\usepackage{times}
\usepackage{amsmath,mathtools}
\usepackage{amsthm}	
\usepackage{amssymb}
\usepackage{enumerate}
\usepackage{booktabs}
\usepackage{tabularx}
\usepackage{array}
\usepackage{bm}
\usepackage{url}
\usepackage{enumitem}
\usepackage{stackengine}
\usepackage{multirow}
\stackMath
\usepackage{lipsum}
\usepackage{subcaption}
\usepackage[numbers,sort,compress]{natbib}
\usepackage{tabularx}
\usepackage{nccmath}

\usepackage[pagebackref=true,breaklinks=true,letterpaper=true,colorlinks,bookmarks=false]{hyperref}

\newcommand{\PBS}[1]{\let\temp=\\#1\let\\=\temp}
\newcommand{\RBS}{\let\\=\tabularnewline}

\newcommand{\modelname}{Deep Bingham Network}
\newcommand{\modelnames}{Deep Bingham Networks}
\newcommand{\modelabbrv}{DBN}
\newcommand{\modelabbrvs}{DBNs}

\newcommand{\x}{\mathbf{x}}

\newcommand{\B}{\mathcal{B}}
\newcommand{\Img}{\mathbf{X}}
\newcommand{\ImgSet}{\mathcal{X}}

\newcommand{\Eye}{\mathbf{I}}
\newcommand{\X}{\mathbf{X}}
\newcommand{\q}{\mathbf{q}}

\newcommand{\qc}{\mathbf{\bar{q}}}
\newcommand{\tb}{\mathbf{t}} 
\newcommand{\Hamil}{{\mathbb{H}}} 
\newcommand{\pars}{{\mathbf{\Gamma}}}

\newcommand{\V}{\mathbf{V}}
\newcommand{\Lo}{{\cal L}}

\newcommand{\Sp}{\mathbb{S}}
\newcommand{\R}{\mathbb{R}}

\DeclareMathOperator*{\argmin}{arg\min}
\DeclareMathOperator*{\argmax}{arg\max}

\newcommand{\etal}{et al.}
\newcommand{\ie}{i.e. }
\newcommand{\eg}{e.g. }

\usepackage[capitalise]{cleveref}

\newcommand{\insertimageC}[5]{ % scale, filename, caption, label, location
\begin{figure}[#5]
\centering
\includegraphics[width=#1\linewidth, clip=true]{figures/#2}
\caption{#3}
\label{#4}
\end{figure}
}

\newcommand{\insertimageStar}[5]{ % scale, filename, caption, label, location
\begin{figure*}[#5]
\centering
\includegraphics[width=#1\linewidth, clip=true]{figures/#2}
\caption{#3}
\label{#4}
\end{figure*}
}

\begin{document}

\title{Deep Bingham Networks}
\subtitle{Dealing with Uncertainty and Ambiguity in Pose Estimation}

\author{Haowen Deng\textsuperscript{1,2} \and Mai Bui\textsuperscript{1} \and Nassir Navab\textsuperscript{1} \and Leonidas Guibas\textsuperscript{3} \and Slobodan Ilic\textsuperscript{2} \and Tolga Birdal\textsuperscript{3}}

\institute{H. Deng* \at
              \email{haowen.deng@tum.de}           %  \\
           \and
           M. Bui* \at
              \email{mai.bui@tum.de}           %  \\
           \and
           N. Navab \at
            \email{nassir.navab@tum.de}           %  \\
           \and
           L. Guibas \at
              \email{guibas@cs.stanford.edu}
              \and
           S. Ilic \at
              \email{slobodan.ilic@tum.de}
              \and
           T. Birdal [0000-0001-7915-7964] \at
              \email{t.birdal@stanford.edu} 
           \and
           1. Informatics at Technische Universität München, Munich, Germany
           \and \\
           2. Corporate Technology Siemens AG, Munich, Germany
           \and \\
           3. Computer Science Department, Stanford University, CA USA
           \and \\
           * shared first authorship
}
\date{}

\maketitle

\begin{figure}
    \centering
    \includegraphics[width=\linewidth]{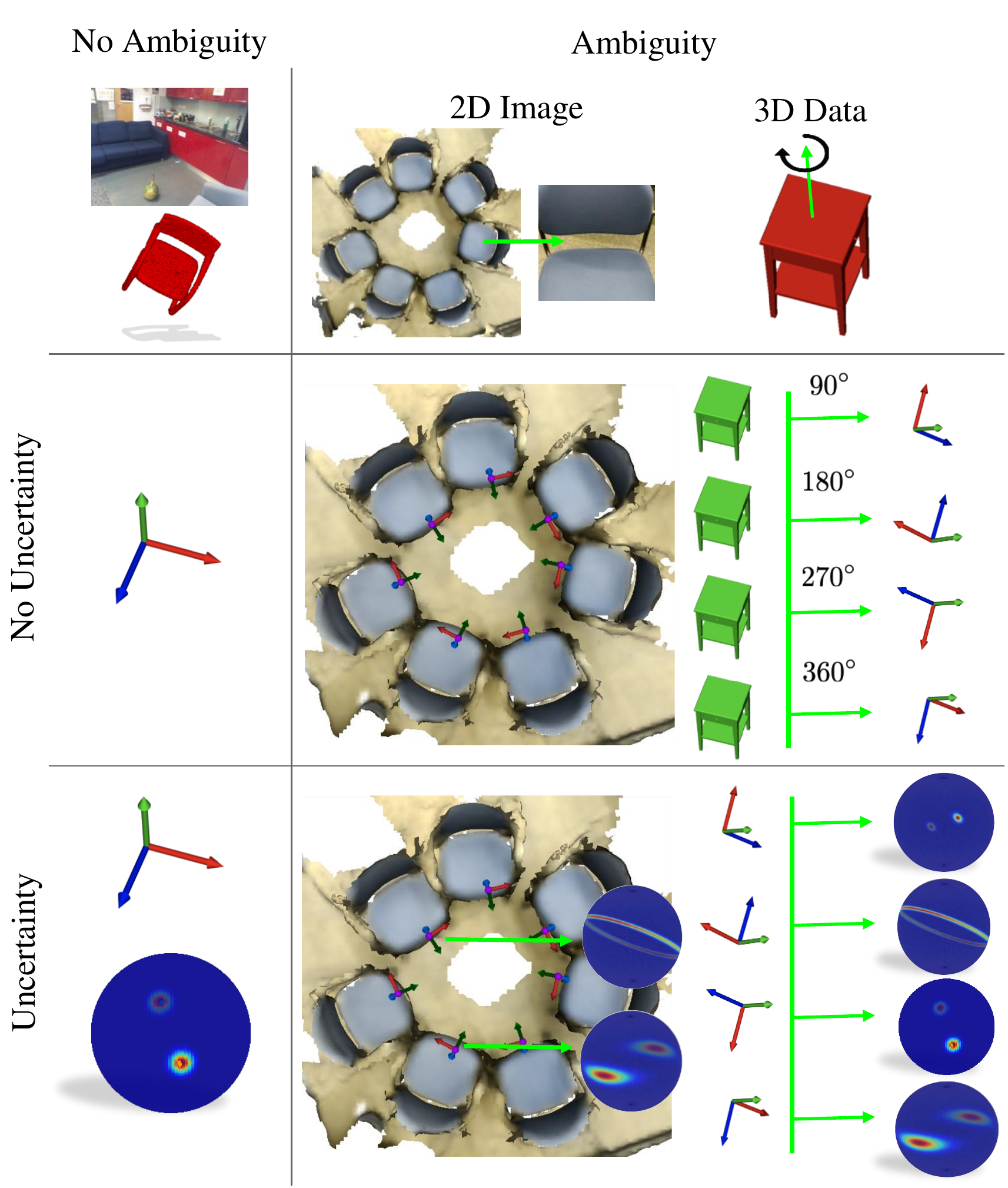}
    \caption{DBM as a generic framework, addresses the following problems arising in pose prediction methods: Estimating 1) a single pose from non-ambiguous input data, 2) multiple pose hypotheses in case of ambiguities; and in addition, associating uncertainty both 3) to a single pose and 4) to all of the hypotheses in the multi-modal predictions.}
    \label{fig:teaser}
\end{figure}
\begin{abstract}
In this work, we introduce \emph{Deep Bingham Networks (DBN)},  a generic framework that can naturally handle pose-related uncertainties and ambiguities arising in almost all real life applications concerning 3D data. While existing works strive to find a single solution to the pose estimation problem, we make peace with the ambiguities causing high uncertainty around which solutions to identify as the best. Instead, we report a \emph{family of poses} which capture the nature of the solution space. DBN extends the state of the art direct pose regression networks by (i) a multi-hypotheses prediction head which can yield different distribution modes; and (ii) novel loss functions that benefit from Bingham distributions on rotations. This way, DBN can work both in unambiguous cases providing uncertainty information, and in ambiguous scenes where an uncertainty per mode is desired. On a technical front, our network regresses continuous \emph{Bingham mixture models} and is applicable to both 2D data such as images and to 3D data such as point clouds. We proposed new training strategies so as to avoid mode or posterior collapse during training and to improve numerical stability. Our methods are thoroughly tested on two different applications exploiting two different modalities: (i) 6D camera relocalization from images; and (ii) object pose estimation from 3D point clouds, demonstrating decent advantages over the state of the art. For the former we contributed our own dataset composed of five indoor scenes where it is unavoidable to capture images corresponding to views that are hard to uniquely identify. For the latter we achieve the top results especially for symmetric objects of ModelNet dataset~\cite{wu20153d}. The code and dataset accompanying this paper is provided under \url{multimodal3dvision.github.io}.
\end{abstract}

\vspace{-6mm}
\section{Introduction}
\label{sec:intro}
A majority of tasks in computer vision can be interpreted as scene understanding problems conditioned on either 2D image or 3D scan modalities. Usually, these scenes are digitizations of our man made environments composed of \emph{objects}. Hence, a fundamental piece of this perception problem is \emph{pose estimation}, \ie figuring out how these objects are positioned and oriented in 3D space. A \emph{rigid transformation} is a six degrees of freedom (6-DoF) entity explaining the pose either of an acquisition device (e.g. Lidar or camera) or an object enclosed within the captured data. Solving for the former is known as \emph{camera relocalization}, while the latter is related to \emph{3D object pose estimation}. 
Both of these are now key technologies in enabling a multitude of applications such as augmented reality, autonomous driving, human computer interaction and robot guidance, thanks to their extensive integration in simultaneous localization and mapping (SLAM)~\cite{durrant2006simultaneous,salas2013slam++,cadena2016past}, structure from motion (SfM)~\cite{ullman1979interpretation,schonberger2016structure}, metrology~\cite{birdal2016online}, visual localization~\cite{shotton2013scene,piasco2018survey} and 3D object detection~\cite{qi2019deep,zhou2018voxelnet}.

A myriad of papers have worked on finding the \emph{unique} solution to the pose estimation problem~\cite{horaud1989analytic,zeisl2015camera,sattler2015hyperpoints,hinterstoisser2012model,zakharov2019dpod,qi2019deep}: a pose per view/scan. However, this trend is now witnessing a fundamental challenge. A recent school of thought has begun to point out that for our highly complex and \emph{ambiguous} real environments, obtaining a single solution \ie the \emph{correct pose}, is simply not sufficient. For example, an image of a scene with repeating structures can look similar even though the location and orientation of the capture device is drastically different. Likewise, objects with rotational symmetries lead to very similar point clouds when scanned from different viewpoints, say with a laser scanner. These observations have led to a paradigm shift that has opened a multitude of research directions focusing on these issues. Instead of estimating a single solution, methods now propose to predict a range of solutions providing multiple pose hypotheses~\cite{manhardt2019explaining}, solutions that can associate \emph{uncertainties} to their predictions~\cite{kendall2016modelling,manhardt2019explaining} or even solutions in the form of full probability distributions~\cite{birdal2018bayesian, arun2018probabilistic,birdal2019probabilistic}.

In this paper, we propose a generic data driven pose estimation algorithm that can handle non-ambiguous as well ambiguous input data and is able to infer multiple solutions with their associated uncertainties. Specifically, depending on the input data given, we can estimate 1) a single pose, 2) multiple pose hypotheses, and in addition associating both 3) a single pose and 4) all hypotheses with a measure of uncertainty in the prediction. \Cref{fig:teaser} summarizes the problems we address in this work. We further propose to capture the multiple plausible solutions of an ambiguous input in the form of a \emph{continuous multimodal distribution} on the Riemannian manifold of poses, while explaining uncertainties by the entropy of the underlying distributions. In particular, we model rotations by a mixture of anisotropic Bingham distributions~\cite{Bingham1974} that are well suited to capture the nature of quaternion parameterizations. We handle translations similarly using Gaussian distributions well suited to capture the variability in Euclidean spaces. 
To be more specific we extend our previous work on camera relocalization in ambiguous scenes~\cite{bui2020eccv} including object pose estimation in ambiguous 3D input.
We begin by explaining our unimodal Bingham distribution based network, termed as \emph{UBN} which can predict a single pose hypothesis and assign a measure of uncertainty to the prediction, but lacks the ability to model ambiguities. We then architect a multi-hypotheses prediction network similar to the one proposed by \cite{rupprecht2017learning,manhardt2019explaining}, termed as \emph{mixture Bingham networks} (MBN). This multi-headed network yields particle predictions spread across the posterior in order to capture different modes. Unlike~\cite{rupprecht2017learning}, we additionally predict the mixture weights and variances, similar to \emph{mixture density networks} (MDN), anchored on each mode resulting in a fully continuous Bingham mixture distribution. With a carefully designed training scheme, we largely alleviate issues such as the mode collapse attributed to MDNs, however, without resorting to a full particle scheme like~\cite{rupprecht2017learning}. We propose to train our networks with a multi-task loss that drives the network to a good optimum between capturing ambiguities and pose prediction itself. We extensively evaluated our methods on two fundamental applications -- \emph{6D camera relocalization} and \emph{object pose estimation from point clouds}. We obtained superior results in comparison to the state-of-the-art especially when the data is inherently ambiguous. 
Our method is flexible in the sense that it can be used with a wide variety of backbone architectures, both for 2D images and for 3D data.

Having a continuous distribution of plausible solutions at hand is useful on multiple fronts: It allows for 1) the estimation of uncertainty~\cite{birdal2018bayesian} and provides a reliable confidence measure, 2) a direct use of the sampled solution space to characterize the 3D object symmetries or configuration of data acquisition, and 3) determining the best solution not through a naive conditional averaging but a scheme that is aware of multiple weighted modes. Note that while in 3D applications, the objects can posses many types of symmetries, unlike~\cite{pitteri2019object,corona2018pose} we avoid making a distinction and rather try to capture this nuance in the multimodal predictions, without explicit supervision.
In conjunction with our earlier work~\cite{bui2020eccv}, our contributions involve:
\begin{enumerate}[noitemsep]
	\item We provide a general framework for continuously modelling conditional density functions on quaternions using Bingham distributions, while explaining the translational uncertainty with multi-modal Gaussians when applicable. Both unimodal and multimodal models are proposed in our work and are extensively evaluated.
	\item For this purpose, we devise novel ways of tailoring neural networks that fit our framework and propose effective multi-task training schemes that are well suited to the complex and non-convex posteriors we are aiming to predict. As a result, we enable an efficient optimization of the necessary distribution parameters while avoiding problems such as mode collapse and numeric instabilities existing in original Mixture Density Networks.
	\item We exhaustively evaluate our methods on two fundamental problems where quantifying the pose is essential: camera relocalization and object pose estimation. We validate our approach, showing that uncertainties captured by our network correlate with the predicted rotation errors. Further, we show that ambiguities could be well handled by our deeply learned Bingham mixture model, both with regard to the quality of the single best prediction as well as the ability of capturing multiple ambiguous modes.
\end{enumerate}

\section{Related Work}
\label{sec:related}
Each of the problems summarized in Fig. \ref{fig:teaser} has posed an ongoing research question. In the following we will briefly outline recent findings for each of these categories.

\paragraph{Single Pose Estimation} Pose estimation is a widely studied topic due to its fundamentality in many vision-based systems, such as CAD model pose estimation from images~\cite{kehl2017ssd, zakharov20173d,birdal2017cad,kanezaki2018rotationnet,birdal2015point}, object pose estimation from point clouds~\cite{qi2019deep,zhao2020quaternion}, camera pose estimation~\cite{kendall2017geometric,kendall2015posenet, kendall2015modelling, brachmann2018learning, bui2019iccw, brachmann2017dsac,bui2018bmvc,feng20166d,massiceti2017random} or pairwise pose alignment~\cite{deng2018ppfnet, deng2019direct,wang2019deep}. Apart from the correspondence based methods used in those applications~\cite{zakharov2019dpod,zeng20173dmatch,deng2018ppf}, direct regression methods from a single input instance are becoming more and more popular due to their simplicity and fast deployment~\cite{kendall2015posenet, xiang2018posecnn, deng2019direct}. Particularly, with the advent of deep learning, powerful feature extraction methods from either images~\cite{he2016deep,deng2009imagenet} or 3D data~\cite{qi2017pointnet, qi2017pointnet++} have emerged that can pave the way to more accurate direct pose regression. Most related to our work, Kendall~\etal~\cite{kendall2015posenet} for example was the first to adopt a convolutional neural network to regress the 6D camera location and orientation from a single RGB image. Similarly, on 3D data, Deng~\etal~\cite{deng2019direct} learned to predict the relative poses between partial scans by regressing the rotations from pairs of local patches. 
Dealing with ambiguities in the context of pose estimation has so far often been handled by prior knowledge of object symmetries. The pose estimation network of Pitteri~\etal~\cite{pitteri2019object} explicitly considered axis-symmetric objects whose pose cannot be uniquely determined. Likewise, Corona~\etal~\cite{corona2018pose} addressed general rotational symmetries. All of these works require extensive knowledge about the object and cannot be extended to the scenario of localizing against a scene without having a 3D model. Note that they also cannot handle the case of self-symmetry and~\cite{corona2018pose} additionally requires a dataset of symmetry-labeled objects, an assumption unlikely to be fulfilled in real applications.

\paragraph{Dealing with Uncertainty}
The above mentioned methods have shown promising results. However, so far most methods neglect that typical CNNs~\cite{Simonyan14c,he2016deep} are over-confident in their predictions~\cite{guo2017calibration,zolfaghari2019learning}. Moreover, these networks tend to approximate the conditional averages of the target data~\cite{bishop1994mixture}. These undesired properties render the immediate outputs of those networks unsuitable for the quantification of calibrated uncertainty, i.e.\ all predictions are assumed to be equally correct~\cite{deng2019direct}. As a result no information of uncertainty can be provided to indicate how good or bad the predictions are.
Initial attempts that address these issues and aim to capture the uncertainty of camera relocalization methods involved the use of random forests~\cite{breiman2001random}. Valentin~\etal~\cite{valentin2015exploiting} stored components of a Gaussian Mixture Model at the leaves of a scene coordinate regression forest~\cite{shotton2013scene}. The modes are obtained via a mean shift procedure on the scene coordinate samples, and the covariance is explained by a 3D Gaussian. A similar approach later considered the uncertainty in object coordinate labels~\cite{brachmann2016uncertainty}. A shortcoming of both of these approaches is the requirement of hand crafted depth features to train the regression forest. Moreover, their uncertainty is on the correspondences and not on the final camera pose. As a result a costly RANSAC~\cite{fischler1981random} is required to propagate the uncertainty in the leaves to the camera pose.

In comparison, probabilistic methods can provide the means to directly capture the uncertainty~\cite{berger2013statistical} in the instance of interest. An initial attempt to capture the variability in the predictions has incorporated Monte Carlo Sampling into neural networks by activating the dropout layers commonly used in neural networks~\cite{gal2016dropout}. For instance, Kendall and Ci-polla \cite{kendall2015modelling} augmented PoseNet \cite{kendall2015posenet} with uncertainty by sampling the posterior to approximate probabilistic inference. In comparison, Mixture Density Networks~\cite{bishop1994mixture} directly learn to predict parameters of a Gaussian mixture distribution by using a neural network which in turn can be used to infer the uncertainty in a prediction based on the variance of the predicted distribution. Yet this method suffers from problems like mode collapse and numeric instability, and Gaussian distributions are not ideal for modeling directional data.  
VidLoc \cite{clark2017vidloc} adapted MDNs \cite{bishop1994mixture} to model and predict uncertainty for the 6D relocalization problem. Besides the reported issues of MDNs, VidLoc incorrectly modeled the rotation parameters using Gaussian distributions and lacked the demonstrations of uncertainty on rotations.
Prokudin~\etal~\cite{prokudin2018deep} replaced the Gaussian distribution with von Mises distribution to enable estimation of a continuous probability space applied to head pose orientations. However, only poses aligned with certain axis are able to be modeled by 2D von Mises distribution. The Bingham distribution~\cite{Bingham1974}, on the other hand, is found to be a good way of analyzing quaternion distributions in a full rotation space. Glover~\etal\cite{glover2012monte, glover2014} estimated the parameters of a Bingham distribution via \textit{Sample Consensus}. The closest to our work has been presented by Gilitschenski~\etal~\cite{Gilitschenski2020Deep} and proposes end-to-end orientation learning by incorporating the Bingham Distribution for object pose estimation from 2D images. However, the method does not yet provide any means of dealing with problems such as mode collapse commonly known to arise in MDNs.

\paragraph{Multiple Hypotheses Prediction}
In general, ambiguities arise due to the existence of multiple legit solutions. For example, in a 3D object pose estimation scenario, it can be caused by an object's rotational symmetries~\cite{corona2018pose,pitteri2019object}, or in a relocalization scenario, identical views acquired by cameras under different poses~\cite{kendall2017uncertainties}.
% Ambiguity is another huge obstacle faced by many researchers. 
Many other prior works targeting ambiguities derive from the field of future prediction~\cite{luc2017predicting, liu2018future}. ~\cite{guzman2012multiple, dey2015predicting} proposed to generate multiple outputs as possible choices, and a \textit{winner takes all (WTA)} strategy was proposed~\cite{guzman2012multiple} and later widely adopted in other applications such as semantic segmentation~\cite{luc2017predicting}. Rupprecht~\etal~\cite{rupprecht2017learning} provided a better way to understand the benefits of this branch of methods with a mathematical formulation, and a relaxation term that was introduced to WTA to facilitate convergence. In these literature, only discrete outputs are considered instead of a continuous space. To close the gap,  Makansi \etal~\cite{makansi2019overcoming} learned to fit parameters of a Gaussian mixture model to the generated point hypotheses in a two-stage training scheme with a variant of WTA loss. 
In pose estimation, to deal with rotational symmetries and self-occlusion symmetries from visual data, Manhardt~\etal~\cite{manhardt2019explaining} generate multiple quaternions as hypotheses for 6D pose estimation.

\paragraph{Dealing with Uncertainty and Ambiguity} Few works have yet attempted to capture both multiple solutions and uncertainty prediction in the context of pose estimation. Monte Carlo sampling for example has been used to create multiple pose predictions~\cite{kendall2016modelling}. Unfortunately even for moderate dimensions these methods still face difficulties in capturing multiple modes. In theory these methods can generate discrete samples from the multi-modal distributions. In practice, as we will demonstrate, the Monte Carlo scheme tends to draw samples around a single mode. This method also suffers from the large errors associated to PoseNet~\cite{kendall2015posenet} itself and can not provide a measure of uncertainty for each pose hypothesis. Manhardt~\etal~\cite{manhardt2019explaining} infer a measure of uncertainty from generated multiple quaternion predictions, however not for each pose hypothesis either. Further, methods predicting a mixture of distributions in theory can capture multiple predictions. Prokudin~ \etal~\cite{prokudin2018deep} for instance learn a variational auto-encoder~\cite{kingma2013auto} to approximate the posterior of $SO(2)$ modeled by von Mises mixtures~\cite{mardia2009directional} and Gilitschenski~\etal~\cite{Gilitschenski2020Deep} show that learned mixture models can aid in handling rotational ambiguities. These approaches, however, do not yet provide any measure of dealing with mode collapse that commonly arises in these type of methods, and therefore are not yet fully able to capture multiple distinct pose predictions.

In comparison our work leverages the best properties from MDN~\cite{bishop1994mixture}, WTA~\cite{rupprecht2017learning, manhardt2019explaining} and Bingham distribution~\cite{Bingham1974} to avoid problems such as mode collapse. Each unimdoal Bingham distribution is treated as a single hypothesis and we aim to capture ambiguities via multiple Bingham distributions predicted by the network, without full modeling of the object symmetries or repeated structures. Eventually, ambiguities can be explained by the modes of a Bingham mixture model while the uncertainty is captured in the concentration parameters or in the \emph{entropy}. Furthermore, we extensively evaluate our method on two applications, namely camera localization from 2D images and object pose estimation from point clouds.

\section{The Bingham Distribution}
We now introduce the mathematical concepts our work is based on, starting with the foundation of our work, the Bingham distribution. The Bingham distribution~\cite{Bingham1974} is an antipodally symmetric probability distribution derived from a zero-mean Gaussian. It is conditioned to lie on $\Sp^{d-1}$ and its probability density function $\B : \Sp^{d-1} \rightarrow \R$ is computed as follows:

\begin{ceqn}
	\begin{align}
	\label{eq:bingham}
	\B(\x; \mathbf{\Lambda}, \V) & = (1/F) \exp(\x^T\V\mathbf{\Lambda}\V^T\x) 
	\\&= (1/F) \exp\big(\sum\nolimits_{i=1}^d \lambda_i(\mathbf{v}_i^T \x)^2 \big)
	\end{align}
\end{ceqn}
where $\V \in \R^{d\times d}$ is an orthogonal matrix $(\V\V^T = \V^T\V = \Eye_{d\times d})$ describing the orientation, $\mathbf{\Lambda} \in \R^{d\times d}$ is called the \textit{concentration matrix} and is constrained such that $0 \geq \lambda_1 \geq \cdots \geq \lambda_{d-1}$: 

\begin{ceqn}
	\begin{align}
	    \mathbf{\Lambda} = \text{diag}\left(\,[\, 0, \lambda_1, \lambda_2, \ldots, \lambda_{d-1}\,]\,\right)
	\end{align}
\end{ceqn}

It is easy to show that adding a multiple of the identity matrix $\Eye_{d\times d}$ to $\V$ does not change the distribution~\cite{Bingham1974}. Thus, we conveniently force the first entry of $\mathbf{\Lambda}$ to be zero. Moreover, since it is possible to swap columns of $\mathbf{\Lambda}$, we can build $\V$ in a sorted fashion. This allows us to obtain \textit{the mode} very easily by taking the first column of $\V$. Due to its antipodally symmetric nature, the mean of the distribution is always zero.
$F$ in~\cref{eq:bingham} denotes the \textit{the normalization constant} dependent only on $\mathbf{\Lambda}$ and is of the form:

\begin{ceqn}
	\begin{equation}
	\label{eq:F}
	F \triangleq |S_{d-1}| \cdot {}_{1}F_1 \Big({1}/{2},\,{d}/{2},\, \mathbf{\Lambda}\Big),
	\end{equation}
\end{ceqn}
where $|S_{d-1}|$ is the surface area of the $d$-sphere and ${}_{1}F_1$ is a confluent hypergeometric function of matrix argument~\cite{carl1995}.

\paragraph{Relationship to quaternions}
The antipodal symmetry of the probability density function $\B(\cdot)$ makes it amenable to explain the topology of quaternions, i. e., $\B(\x; \cdot) = \B(-\x; \cdot)$ holds for all $\x \in \Sp^{d-1}$. 
In 4D when $\lambda_1=\lambda_2=\lambda_3$, one can write $\mathbf{\Lambda}=\text{diag}([1,0,0,0])$. In this case, the Bingham density relates to the dot product of two quaternions $\q_1\triangleq\x$ and the mode of the distribution, say $\qc_2$. This induces a metric of the form
\begin{ceqn}
    \begin{equation}
        d_{\text{bingham}}=d(\q_1,\q_2)=(\q_1 \cdot \qc_2)^2 = \text{cos}(\theta/2)^2,
 	\end{equation}
\end{ceqn}
that is closely related to the true Riemannian distance~\cite{birdal2018bayesian} given two rotation matrices $\mathbf{R}_1$ and $\mathbf{R}_2$
\begin{ceqn}
    \begin{align}
        d_{\text{riemann}}=\|\text{log}(\mathbf{R}_1 \mathbf{R}_2^T)\| &\triangleq 2 \text{arccos}(|\q_1 \qc_2|)\\
        &\triangleq 2\text{arccos}(\sqrt{d_{\text{bingham}}}).
 	\end{align}
\end{ceqn}
Bingham distributions have been extensively used to represent distributions on quaternions ($\Hamil_1$)~\cite{glover2012monte,kurz2013recursive,glover2014,birdal2018bayesian,birdal2020measure}.

\paragraph{Relationship to other representations}
Note that geometric~\cite{barfoot2014associating} or measure theoretic~\cite{falorsi2019reparameterizing}, there are multitudes of ways of defining probability distributions on the Lie group of 6D rigid transformations~\cite{haarbach2018survey}. A naive choice would be to define Gaussian distribution on the Rodrigues vector (or exponential coordinates)~\cite{murray1994} where the geodesics are straight lines~\cite{morawiec1996rodrigues}. However, as our purpose is not tracking but direct regression, in this work we favor quaternions as continuous and minimally redundant parameterizations without singularities~\cite{grassia1998,busam2017camera} and use the Bingham distribution that is well suited to their topology. We handle the redundancy $\mathbf{q}\equiv-\mathbf{q}$ by mapping all the rotations to the northern hemisphere. 

\subsection{Constructing Orientation Matrices $\V$}
\label{sec:V}
%\paragraph{Construction of orthogonal matrix $\V$}
Unlike Gaussian distributions whose covariance is aligned with the standard basis, constructing a Bingham distribution requires a local frame estimation to establish the orientation matrix $\V$. While the first component of this matrix is the \emph{mode} as explained above, it is not clear how the other components should be computed. Additionally, ensuring this matrix is orthonormal requires care as adding a regularization term such as $||\V^\top\V - \Eye||_F$ during optimization cannot guarantee a valid orthonormal matrix. In this work, we investigate three different ways to construct $\V$:
\begin{enumerate}
    \item \textbf{Gram-Schmidt process}\, A straightforward way is to first estimate an unconstrained Euclidean matrix $\mathbf{M} \in \R^{d \times d}$ and then ortho-normalize it into $\V$ via Gram-Schmidt (GS) process. In this case, the column vectors $\mathbf{v_i}$ of $\V$ are computed from the column vectors $\mathbf{m_i}$ as follows
\begin{ceqn}
	\begin{equation}
			\mathbf{\hat{v}_i} = \mathbf{m}_i - \sum_{k=1}^{i-1} \langle \mathbf{v}_k,\mathbf{m}_i \rangle \cdot \mathbf{v}_k~,\text{where}~
			\mathbf{v}_i = \frac{\mathbf{\hat{v}}_i}{\|\mathbf{\hat{v}}_i\|}.
	\label{eq:V_gs}
	\end{equation}
	\end{ceqn}
	This GS procedure requires prediction of 16 values, an over-parametrization of the degrees of freedom in $\V$. In the following, we refer to this process as \textit{Gram-Schmidt (GS) Strategy}.\\
	\item \textbf{Matrix representation}\, 
	To use the minimal degrees of freedom, an elegant way proposed by Birdal~\cite{birdal2018bayesian} is to estimate the mode $\q \in \Hamil_1$ and subsequently find a set of vectors orthonormal to $\q$. Fortunately, the quaternions can be linearly represented by matrices. In other words, there exists an injective homomorphisms from $\Hamil_1$ to the matrix ring $\text{M}(4, \R)$. The result is a frame bundle $\Hamil_1 \rightarrow \mathcal{F}\Hamil_1$ composed of four unit basis vectors: the mode and its orthonormals:
\begin{ceqn}
\begin{align}
\label{eq:V}
\V(\q) \triangleq 
\begin{bmatrix}
q_1 & -q_2 				& -q_3 				&  \phantom{-}q_4 \\
q_2 & \phantom{-}q_1 	& \phantom{-}q_4 	&  \phantom{-}q_3\\
q_3 & -q_4 				& \phantom{-}q_1 	&  -q_2\\
q_4 & \phantom{-}q_3 	& -q_2				&   -q_1
\end{bmatrix}.
\end{align}
\end{ceqn}
It is easy to verify that the matrix valued function $\V(\q)$ is orthonormal for every $\q \in \Hamil_1$. 
$\V(\q)$ further gives a convenient way to represent quaternions as matrices paving the way to linear operations, such as quaternion multiplication or orthonormalization without the Gram-Schmidt. We refer this one as \textit{Birdal Strategy}.\\

\item \textbf{Cayley transformation}\,
Utilizing the Cayley transform, which describes a mapping from skew-symmetric matrices to special orthogonal matrices, we propose a third way to construct $\V$:  Given the mode $\q$ (not necessarily with unit norm), we compute $\V$ as:
\begin{ceqn}
	\begin{equation}
		\V = (\Eye_{d \times d} - \mathbf{S})^{-1} (\Eye_{d \times d} + \mathbf{S}),
	\end{equation}
\end{ceqn}
where $\Eye_{d \times d}$ is the identity matrix and
\begin{ceqn}
\begin{align}
\label{eq:S}
\mathbf{S}(\q) \triangleq 
\begin{bmatrix}
\phantom{-}0       & -q_1 	        & q_4 	        &  -q_3 \\
\phantom{-}q_1     & \phantom{-}0 	& q_3 	        &  \phantom{-}q_2\\
-q_4               & -q_3 	        & \phantom{-}0 	&  -q_1\\
\phantom{-}q_3     & -q_2 	        & q_1	        &  \phantom{-}0
\end{bmatrix}
\end{align}
\end{ceqn}
is a skew-symmetric matrix parameterized by $\q$. Similar to \emph{Birdal} this allows us to only estimate four values and even removes the need of normalization to obtain a valid quaternion during optimization. We term this construction as \textit{Cayley Strategy}.
\end{enumerate}

Note that, for \textit{Birdal} and \textit{Cayley} strategies, a reduced number of predictions (4) suffice to yield $\V$ compared to \textit{Gram-Schmidt} (16). We will show later in our experiments that the former two also demonstrate better performance.

\insertimageStar{1}{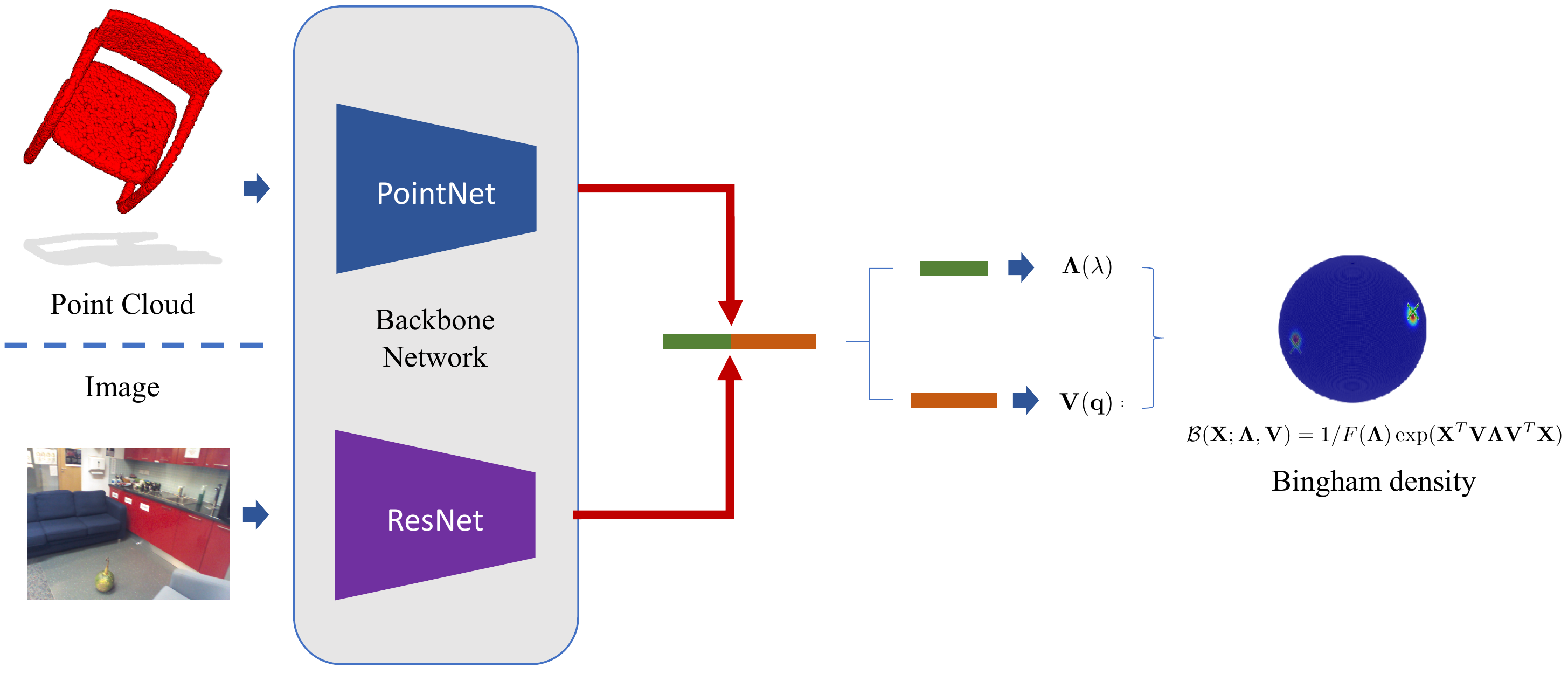}{The pipeline for Unimodal Bingham Network. The input data is processed by an adequate backbone network (PointNet for a rotated point cloud and ResNet for a 2D image here) to output a 7-$d$ vector from the last layer, which is later used to form $\mathbf{\Lambda}$ and $\V$ of a Bingham distribution. }{fig:unimodal_bingham_network}{t!}
\section{\modelnames~(\modelabbrvs)}
\label{sec:DBN}
The Bingham distribution establishes the foundation of modelling orientations while providing the means for uncertainty estimation as well. We now further elaborate on how such a distribution can be integrated into a neural network to provide both the means for uncertainty estimation as well as to handle ambiguity issues in pose estimation problems, without sacrificing the accuracy of the single prediction For this aim, we adopt deep neural networks and predict the underlying posterior distribution of the target pose in an end-to-end style. We consider the situation where we observe an input image $\Img\in \R^{W\times H \times 3}$ or a point cloud $\Img\in \R^{N \times 3}$ and assume the availability of a predictor function $\V\triangleq \V_{\pars}(\Img)$ parameterized by $\pars=\{\pars_i\}$. $\V(\cdot)$ is an orthogonal orientation matrix computed using any of the  strategies introduced in~\cref{sec:V}. Note that predicting entities that are non-Euclidean easily generalizes to prediction of Euclidean quantities such as translations e.g. $\tb\in\R^3$ when the constraints are removed.
To this end, we investigate two models:\\

\vspace{-1mm}\noindent\textbf{Unimodal Bingham Network (UBN)} models the pose by a single Bingham distribution. In addition to computing a single best prediction, the entropy of the resulting distribution can be used as a measure of the uncertainty. An overview of this model is shown in Fig. \ref{fig:unimodal_bingham_network}.

\vspace{1mm}\noindent\textbf{Multimodal Bingham Network (MBN)} predicts a multimodal Bingham
distribution, as shown in Fig. \ref{fig:multimodal_bingham_network}. It extends the advantage of UBN with extra capability of capturing different modes lying in the data, thus dissolving ambiguities.
\vspace{0.5mm}

Note the definitions of our \modelname~do not include any specific network architectures, i.e. they are agnostic of the backbone networks and can be combined with any existing networks other than the reported ones in our paper. 

\subsection{Unimodal Bingham Network (UBN)}
\label{sec:uncertainty}
We now start with describing our models in more detail, beginning with our Unimodal Bingham Network. UBN takes an observation in the form of either a point cloud $\Img\in\R^{N\times 3}$ or a 2D image $\Img\in\R^{W\times H \times 3}$, and predicts the essential parameters of a unimodal Bingham distribution of the target pose. It describes the orientation of the object (or the camera) of interest, i.e. a single pose prediction with associated measure of uncertainty given by the entropy of the distribution.

The first column of $\V$ represents the correct values of the rotation $\q_i\in\Hamil_1$, admitting a non-ambiguous prediction, hence a posterior of single mode. We use the predicted rotation to set the most likely value (mode) of a Bingham distribution:

\begin{ceqn}
    \begin{align}
        p_{\pars} (\q \,|\, \Img ; \mathbf{\Lambda}) = (1/F)~\text{exp } \big(\q^\top \V \mathbf{\Lambda} \V^\top \q \big),
    \end{align}
\end{ceqn}
and let $\q_i$ differ from this value up to the extent determined by $\mathbf{\Lambda}=\{\lambda_i\}$. For the sake of brevity we use $\V\equiv\V_{\pars}(\Img)$. In this work, we model $\pars$ by a deep neural network.

While for certain applications, fixing $\mathbf{\Lambda}$ can work, in order to capture the variation in the input, it is recommended to adapt $\mathbf{\Lambda}$~\cite{prokudin2018deep}. Thus, we introduce it among the unknowns. To this end we define the function ${\mathbf{\Lambda}}_{\pars}(\Img)$ or in short $\mathbf{\Lambda}$ for computing the concentration values depending on the current input $\X$ and the parameters ${\pars}$. Our final model for the unimodal case reads:

\begin{ceqn}
    \begin{align}
        \label{eq:posterior}
        p_{\pars} (\q \,|\, \Img) & = \frac{\text{exp } \big(\q^\top \V_{\pars}(\Img) \mathbf{\Lambda}_{\pars}(\Img) \V_{\pars}(\Img)^\top \q \big)}{F(\mathbf{\Lambda}_{\pars}(\Img))}
        \\&= \frac{\text{exp } \big(\q^\top \V \mathbf{\Lambda} \V^\top \q \big)}{F(\mathbf{\Lambda})}
    \end{align}
\end{ceqn}
The second row follows from the short-hand notations and is included for
clarity. For the rest of this section, we stick with this simplified notations.

\paragraph{Inferring Bingham Parameters}
To produce a Bingham probability model with a network, we need to propose an end-to-end inference paradigm for $\mathbf{\Lambda}$ and $\V$, conditioned only on the input $\Img$. We need to ensure all values in $\mathbf{\Lambda}$ to be non-positive and in descending order and $\V$ to be an orthogonal matrix.

Assume a backbone feature network is available, the original output is unconstrained and does not satisfy the properties of $\V$ and $\mathbf{\Lambda}$. Further processing of those raw values is necessary. In order to keep the efforts on modifications to the lowest level, we propose only to change the last layer without adapting the remaining part of the network.

Three values are needed in $\mathbf{\Lambda}$.
In order to keep the diagonal values $(\lambda_1, \lambda_2, \lambda_3)$ sorted in \textbf{descending order} and \textbf{non-positive}, we make use of softplus activation and accumulative sum, with an extra negating operation:

\begin{ceqn}
    \begin{align}
        \lambda_1 & = -\phi(o_1)                         \\
        \lambda_2 & = -\phi(o_1) -\phi(o_2)              \\
        \lambda_3 & =  -\phi(o_1) -\phi(o_2) - \phi(o_3).
        \label{eq:lambda}
    \end{align}
\end{ceqn}
$\phi(\cdot)$ is the \textit{softplus} activation function and $o_i (i=1, 2,3)$ are three values taken from the original output of the network.

For constructing $\V$, based on the chosen strategy,  4 or 16 values are needed from the original network output. Also, for \textit{Birdal Strategy}, we need to normalize the four values first to make it a legal quaternion to feed into ~\cref{eq:V}.

The hard-to-compute entity $F$ is shown in \cref{eq:F} to depend solely on $\mathbf{\Lambda}$. To enable fast inference of $F$ and gradient flow from $\mathbf{\Lambda}$ through $F$, we make use of a lookup table that is pre-computed based on a set of predefined range of values in $\mathbf{\Lambda}$~\cite{kume2005saddlepoint,kurz2017directional}.

\paragraph{Entropy as Uncertainty} After obtaining the parameters of the probability function $\B$ of a Bingham distribution, its entropy can be computed.
\begin{ceqn}
    \begin{align}
        E(\B) = \mathrm{log}(F) - \mathbf{\Lambda} \frac{ \nabla F(\mathbf{\Lambda})}{F}
        \label{eq:bingham_entropy}
    \end{align}
\end{ceqn}
Information theory~\cite{jaynes1957information} proves that higher entropy is an indicator of higher uncertainty. 

Following information theory~\cite{jaynes1957information} and similar to~\cite{makansi2019overcoming}, we treat the entropy of the predicted Bingham distribution as a practical measure of uncertainty~\cite{wang2008probability}: the higher entropy the higher uncertainty. For an easier to interpret uncertainty score, we pass the entropy values through a sigmoid function mapping it to the range $(0, 1)$:
\begin{ceqn}
    \begin{align}
        U & = \text{sigmoid}(E(\B))  = \frac{1}{1 + e^{-E(\B)}}
    \end{align}
\end{ceqn}

\insertimageStar{1}{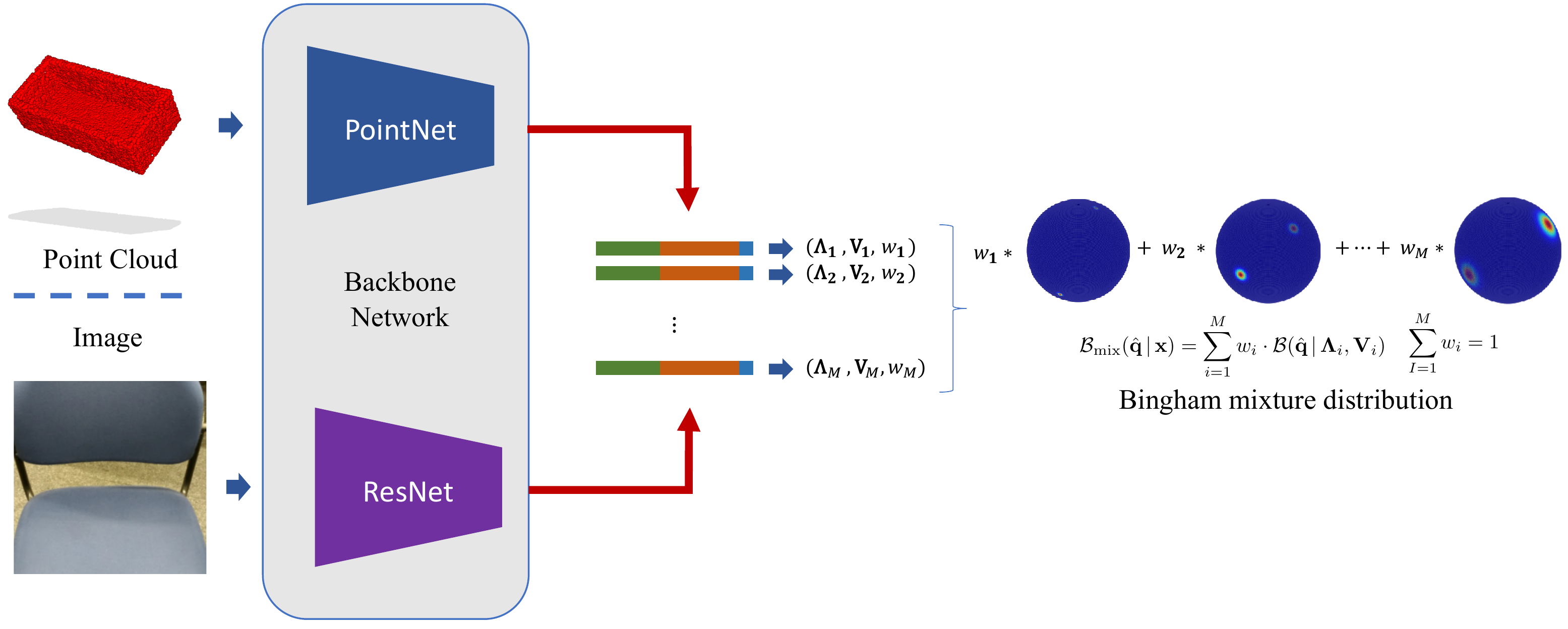}{The pipeline for Multimodal Bingham Network. Same network is used as in UBN, only the last layer is modified to output $M\times(7+1)$ units to form $M$ groups of parameters and weights for different Bingham components. Different modes counting for ambiguity are captured by different Bingham distribution.}{fig:multimodal_bingham_network}{t!}

\paragraph{Bingham Loss} Given a collection of observations $\ImgSet=\{\X_i\}$ and associated rotations $\mathcal{Q}=\{\q_i\}$, we learn the parameters $\pars$ of our unimodal Bingham network by minimizing the negative log-likelihood:

\begin{ceqn}
    \begin{align}
        \Lo (\q\,,\, \B(\mathbf{\Lambda}, \V)) =  {\text{log }F\big({\mathbf{\Lambda}}\big)} - \q^\top \V {\mathbf{\Lambda}} \V^\top \q
    \end{align}
\end{ceqn}

As shown in~\cref{fig:unimodal_bingham_network}, we compose $\V:\V_{\pars}(\X)$ and ${\mathbf{\Lambda}}:{\mathbf{\Lambda}}_{\pars}(\X)$ during training so as to evaluate the density. This maximizes the probability of ground truth $\q_i$ evaluated on the associated Bingham distribution $\B(\mathbf{\Lambda}_i, \V_i)$. In short we obtain the optimal parameters $\pars^\star$ by

\begin{ceqn}
    \begin{align}
        \label{eq:train_single}
        \pars^\star = \argmin_{\pars} \sum_{i=1}^N \Lo ( \q_i, \B(\mathbf{\Lambda}_i, \V_i) \, |\, \pars)
    \end{align}
\end{ceqn}
Note once again that $\mathbf{\Lambda}$ and $\V$ are dependant upon $\X$, thus $\mathbf{\Lambda}_i\equiv \mathbf{\Lambda}_{\pars}(\X_i)$ and $\V_i\equiv\V_{\pars}(
    \mu(\X_i))$.

\subsection{Multimodal Bingham Network (MBN)}
While UBN is able to grant the predictions with uncertainty information, it cannot handle ambiguities such as objects with rotational symmetries, or scenes with identical views under different camera locations. Inspired by MDN~\cite{bishop1994mixture}, we resort to multimodality and propose a Multimodal Bingham Network for predicting multiple Bingham distributions to explain a single observation $\X$, and eventually yielding a Bingham mixture model, as depicted in \cref{fig:multimodal_bingham_network}.

\paragraph{Bingham Mixture Model (BMM)}
A Bingham mixture model can be easily extended from a set of unimodal Bingham models by assigning each one a weight factor and combining them linearly to form a continuous distribution space with multimodality.
Each unimodal \textit{component} captures a \textit{peak} presence of a valid solution. MBN then aims to predict a Bingham Mixture Model (BMM) for any given input $\X$, storing individual component predictions in different \textit{branches}.

We build MBN on top of UBNs. Apart from the parameters $(\mathbf{\Lambda}_i, \V_i)$ for each unimodal component, it also predicts a set of weights $\{w_i\}_{i=1}^{M}$. The probability density function $\B_{\text{mix}}$ can be written as:

\begin{ceqn}
    \begin{align}
        \B_{\text{mix}}(\q\,|\,\X) = \sum_{i=1}^{M} w_i \cdot \B(\q\,|\, \mathbf{\Lambda}_i, \V_i) \,\, : \,\,
        \sum_{i=1}^M w_i & = 1
    \end{align}
\end{ceqn}
$M$ is the number of components.
This way, different poses associated with the same input $\X$ can acquire high probabilities in different components.

\paragraph{Mixture Bingham Loss} Similar to \cref{eq:train_single}, we can define a \textit{Mixture Bingham Loss} as the negative log-likelihood of the mixture distribution model, which can be viewed as a weighted sum of \textit{Bingham losses} per each component.

\begin{ceqn}
    \begin{align}
        \mathcal{L}_{\text{MB}}(\q\,,\, \B_{\text{mix}}) = 
        - \text{log}(\B_{\text{mix}}(\q\,|\,\X) )
    \end{align}
\end{ceqn}
Previous work on Mixture Density Network~\cite{bishop1994mixture,makansi2019overcoming} has shown that directly optimizing all the parameters at the same time can lead to numerical instabilities and problems such as mode collapse might arise. Thus a proper solution to tackle these issues is critical.

\paragraph{RWTA Loss}
\label{sec:wta_loss}

Our current model easily suffers from mode collapse, i.e. it is not able to capture multiple distinct modes well. However, to efficiently handle ambiguities predicting plausible distinct solutions is of importance. With this additional property included into our model, we would be able to capture multiple solutions as well as associated uncertainties for each prediction. We therefore incoporate a "Winner Takes All" training scheme that has been proven to be effective for coping with ambiguities \eg multiple modes~\cite{rupprecht2017learning, manhardt2019explaining}. In WTA, each iteration updates only the branch that generates the closest prediction to the ground truth. This provably leads to the Voronoi tesselation of the output space.

Let $\B_i$ be the $i$-th component of our Bingham Mixture Model. At each training iteration, we select the component giving the best prediction with regard to the current ground truth. We could select the best component by checking the $l_1$ distances of the predicted quaternion $\hat{\q}$ with regard to the given ground truth $\q$~\cite{manhardt2019explaining}.

\begin{ceqn}
    \begin{align}
        i^*  = \argmin_{i} |\q - \hat{\q_i}|
        \label{eq:select_l1}
    \end{align}
\end{ceqn}

An alternative way is to check the probabilities of the ground truth $\q$ on the set of predicted Bingham distributions, and keep the one with highest value.

\begin{ceqn}
    \begin{align}
        i^*  = \argmax_{i} \B_i(\q \,|\, \mathbf{\Lambda}_i, \V_i)
        \label{eq:select_prob}
    \end{align}
\end{ceqn}

Then we optimize $(\mathbf{\Lambda}_i, \V_i)$ of $\B_i$ by minimizing its corresponding Bingham Loss. Rupprecht~\etal~\cite{rupprecht2017learning} find that allowing a small portion of gradients from the sum of losses would help avoid "dead" particles which never get updated because of their bad random initializations, which can be considered as a \textit{relaxed} version of WTA.

\begin{ceqn}
    \begin{align}
        \Lo_{\text{RWTA}} (\q\,,\, \B_{\text{mix}}) & = \sum_{i = 1}^M\pi_i \Lo(\q\,,\, \B_i(\mathbf{\Lambda}_i, \V_i)) \\
        \pi_i                                       & = \begin{cases}
            1 - \epsilon \quad \text{if $i$-th component is selected} \\
            \frac{\epsilon}{M - 1} \quad \text{else}
        \end{cases}
    \end{align}
\end{ceqn}
$\mathcal{L}_{\text{RWTA}}$ is able to guide the multiple unimodal components to cover different modes of the ground truth distribution. Yet, it cannot represent a continuous distribution due to the lack of variance predictions and the mixture weights $\{w_i\}$.

\paragraph{Cross Entropy Loss}
We provide an alternative way to explicitly train the weights for the components in MBN by forming it as a classification problem. 
\begin{ceqn}
    \begin{equation}
        \Lo_{\text{CE}}(w_{\B_{\text{mix}}}) = \sum_{i=1}^{M} -(y_i\cdot\log(w_i) + (1 - y_i) \cdot \log (1 - w_i)),
    \end{equation}
\end{ceqn}
$w_i$ is the predicted weight for the $i-th$ component and $y_i$ is the associated label given by the selection results by either \cref{eq:select_l1} or \cref{eq:select_prob}.

\begin{ceqn}
    \begin{equation}
        y_i = \begin{cases}
            1, & \text{if } i = i^* \\
            0, & \text{otherwise}
        \end{cases}.
    \end{equation}
\end{ceqn}

Based on those loss functions, we propose two independent training schemes:
\begin{itemize}
    \item \textbf{Cross Entropy + RWTA}. Following \cite{bui2020eccv}, it relies on RWTA loss to train each individual components for capturing different modes and Cross Entropy loss to assign proper weights for each one with 
    \begin{ceqn}
        \begin{align}
            \mathcal{L}_{\textnormal{MBN-CE}} = \mathcal{L}_{\textnormal{CE}} + \mathcal{L}_{\textnormal{RWTA}}.
            %\label{eq:loss_mix}
        \end{align}
    \end{ceqn} 
    \item \textbf{Mixture Bingham + RWTA}. It follows the conventional MDN training~\cite{bishop1994mixture} scheme, but imports extra RWTA loss to help guide the training to overcome the exisiting problems resulting in the following loss function
    \begin{ceqn}
        \begin{align}
            \mathcal{L}_{\textnormal{MBN}} = \mathcal{L}_{\textnormal{MB}} + \mathcal{L}_{\textnormal{RWTA}}.
            \label{eq:loss_mix}
        \end{align}
    \end{ceqn}
\end{itemize}

We will show later in our experiments that both schemes could facilitate the training process as well as further improve the performance. To differentiate, we name the MBNs trained with the two schemes as \textit{MBN} and \textit{MBN-CE} respectively. Note that by learning the weights of the mixture model, we can always also find a single best prediction by utilizing the mode associated to the most likely mixture component.

\section{Application to Camera Re-localization}
We first evaluate our method on the task of re-localizing a camera in a given reference scene, before demonstrating our method's performance in predicting the pose of an object from a given input point cloud.

\subsection{Modelling translations}
As a camera's pose is defined by its orientation as well as its position, we predict the rotation by our proposed \modelname. Further, as they reside in the Euclidean space, we use mixture density networks \cite{bishop1994mixture} to incorporate translations. In more detail, for a sample input image $\Img\in \R^{W \times H \times 3}$, we obtain a predicted translation $\hat{\tb} \in\R^{c=3}$ from a neural network with parameters  $\pars$. This prediction is set to the most likely value of a multivariate Gaussian distribution with covariance matrix

\begin{ceqn}
	\begin{align}
	    \mathbf{\Sigma} = \begin{bmatrix}
	    \sigma_1^2 & & \\
	    &  \ddots & \\
	    &  & \sigma_{c}^2
	    \end{bmatrix}_{c \times c},
	\end{align}
	
\end{ceqn}
\noindent where $\boldsymbol{\sigma}^2$ is predicted by our model.
As a result our model for a unimodal Gaussian is defined as:

\begin{ceqn}
	\begin{equation}
	\label{eq:gauss}
		p_{\pars} (\textbf{t} \,|\, \Img) = \frac{\text{exp} (-\frac{1}{2}(\tb - \hat{\tb})^\top \mathbf{\Sigma}^{-1} (\tb - \hat{\tb}))}{(2\pi)^{c/2} |\mathbf{\Sigma}|^{1/2}} ,
	\end{equation}
\end{ceqn}

\noindent where $c=3$ and both $\hat{\tb}$ as well as $\mathbf{\Sigma}$ are trained by minimizing its negative log-likelihood.

Similar to forming a Bingham Mixture Model, we can equally compute a Gaussian Mixture Model with $M$ components and corresponding weights $\{w_i\}$, such that $\sum_{i=1}^{M} w_i = 1$, to obtain a multi-modal solution. Again both $\hat{\tb}$ and $\mathbf{\Sigma}$ as well as $\{w_i\}$ are learned by the network and trained by minimizing the negative log-likelihood of the mixture model. Note that, in this case, the components of $\hat{\tb}$ are assumed to be statistically independent within each distribution component. However, it has been shown that any density function can be approximated up to a certain error by a multivariate Gaussian mixture model with underlying kernel function as defined in \cref{eq:gauss} \cite{bishop1994mixture, mclachlan1988mixture}. In practice we first train our network for the translation and its variance. We then train for the rotation and its distribution parameters, which intuitively, after knowing the position, should be an easier task. Finally we fine-tune the entire network for all distribution parameters.

Similar to our Bingham model we use the entropy of the resulting distribution as a measure of uncertainty:

\begin{ceqn}
	\begin{align}
	E(G) = \frac{c}{2} + \frac{c}{2} \text{log}(2\pi) + \frac{1}{2} \text{log}(|\bm{\Sigma}|),
	\end{align}
\end{ceqn}
respectively, where $c=3$ the dimension of the mean vector of the Gaussian. For a given image we first normalize the entropy values over all pose hypothesis, and finally obtain a measure of (un)certainty for a camera pose hypothesis as the sum of both rotational ($E(B)$, see \cref{eq:bingham_entropy}) and translational ($E(G)$) normalized entropy.  
\label{sec:exp}

\subsection{Implementation Details}
We implement our method in Python using the PyTorch library~\cite{pytorch}. Following the current state-of-the-art direct camera pose regression methods, we use a \textit{ResNet-34}~\cite{he2016deep} as our backbone network architecture, followed by fully-connected layers for rotation and translation, respectively. In the following we use the the strategy of \textit{Birdal} to construct $\V$ and normalize the predicted quaternions during training. Ablation studies on further construction methods are presented in section \ref{sec:ablation}. All models are trained with the ADAM optimizer with an exponential learning rate decay and trained for 300 epochs with a batch size of 20 images.

\subsection{Experimental Setup for 6D Relocalization}
\begin{table*}[t]
	\begin{center}
		\caption{Evaluation in non-ambiguous scenes, displayed is the median rotation and translation error. (Numbers for MapNet on the Cambridge Landmarks dataset are taken from \cite{sattler2019understanding}). BPN depicts Bayesian-PoseNet~\cite{brahmbhatt2018geometry}. \textit{UBN} and \textit{MBN-MB} refer to our unimodal version and mixture model respectively.}
		\resizebox{\textwidth}{!}{
			\begin{tabular}{lccccccc|ccccc}
			    \toprule
				Dataset & \multicolumn{7}{c}{7-Scenes} & \multicolumn{4}{c}{Cambridge Landmarks}\\
				\noalign{\smallskip}
				 $[^\circ~/~\text{m}]$& Chess & Fire & Heads & Office & Pumpkin & Kitchen & Stairs  & Kings & Hospital & Shop & St. Marys & Street\\
				\noalign{\smallskip}
				\midrule
				\noalign{\smallskip}
				PoseNet & $4.48$/$0.13$& $11.3$/$0.27$ & $13.0$/$0.17$ & $5.55$/$0.19$ & $4.75$/$0.26$ & $5.35$/$0.23$ & $12.4$/$0.35$ & $1.04$/$0.88$ & $3.29$/$3.2$ & $3.78$/$0.88$ & $3.32$/$1.57$ & $25.5$/$20.3$\\
				MapNet & $3.25$/$0.08$ & $11.69$/$0.27$ & $13.2$/$0.18$ & $5.15$/$0.17$ & $4.02$/$0.22$ & $4.93$/$0.23$ & $12.08$/$0.3$ & $1.89$/$1.07$ & $3.91$/$1.94$ & $4.22$/$1.49$ & $4.53$/$2.0$ &-\\
				\midrule
				\noalign{\smallskip}
				BPN & $7.24$/$0.37$ & $13.7$/$0.43$ & $12.0$/$0.31$ & $8.04$/$0.48$ & $7.08$/$0.61$ & $7.54$/$0.58$ & $13.1$/ $0.48$ & $4.06$/$1.74$ & $5.12$/$2.57$ & $7.54$/$1.25$ & $8.38$/$2.11$&-\\
				VidLoc& -/$0.18$ & -/$0.26$ & -/$0.14$ & -/$0.26$ & -/$0.36$ & -/$0.31$ & -/$0.26$ & - & - & - & - &-\\
				UBN & $4.97$/$0.1$ & $12.87$/$0.27$ & $14.05$/$0.12$ & $7.52$/$0.2$ & $7.11$/$0.23$ & $8.25$/$0.19$ & $13.1$/$0.28$ & $1.77$/$0.88$ & $3.71$/$1.93$ & $4.74$/$0.8$ & $6.19$/$1.84$ & $24.08$/$16.8$\\
				MBN-MB & $4.35$/$0.1$ & $11.86$/$0.28$ & $12.76$/$0.12$ & $6.55$/$0.19$ & $6.9$/$0.22$ & $8.08$/$0.21$ & $9.98$/$0.31$ & $2.08$/$0.83$ & $3.64$/$2.16$ & $4.93$/$0.92$ & $6.03$/$1.37$ & $36.88$/$9.69$\\
				\bottomrule
			\end{tabular}
		}
 		\label{table:results}
	\end{center}
\end{table*}
When evaluating our method we consider two cases: (1) camera relocalization in non-ambiguous scenes, where our aim is to not only predict the camera pose, but the posterior of both rotation and translation that can be used to associate each pose with a measure of uncertainty; (2) we create a highly ambiguous environment, where similar looking images are captured from very different viewpoints. We show the problems current regression methods suffer from in handling such scenarios and in contrast show the merit of our proposed method. 
\paragraph{Error metrics} 
Given a ground truth camera pose, consisting of a rotation, represented by a quaternion $\q$, and its translation, $\tb$, we evaluate the performance of our models with respect to the accuracy of the predicted camera poses by computing the recall of ours and the baseline models. We consider a camera pose estimate to be correct if both rotation and translation are below a pre-defined threshold and compute the angular error between ground truth, $\q$, and predicted quaternion, $\hat{\q}$, as 

\begin{ceqn}
	\begin{equation}
	    d_q(\q,\hat{\q}) = 2\arccos(|\q \circ \hat{\q}|).
	\end{equation}
\end{ceqn}
For translations we use the norm of the difference between ground truth $\tb$, and predicted translation $\hat{\tb}$: $d_t(\tb,\hat{\tb}) = \|\tb - \hat{\tb}\|_2$ to compute the error in position of the camera.

We obtain a single prediction from our network by taking the weighted mode, the mode of the distribution with highest mixture coefficient. Note that, under ambiguities a best mode is unlikely to exist. In those cases, as long as we can generate a hypothesis that is close to the ground truth, our network is considered successful. For this reason, in addition to the standard metrics and the weighted mode, we will also speak of the so called \textit{Oracle} error, assuming an oracle that is able to choose the best of all predictions: the one closest to the ground truth. In addition, we report the \textit{Self-EMD} (SEMD)~\cite{makansi2019overcoming}, the earth movers distance~\cite{rubner2000earth} of turning a multi-modal distribution into a unimodal one. With this measure we can evaluate the diversity of predictions. We choose the predicted mode, the unimodal distribution of the weighted mode, as reference for this measure. Note that this metric itself does not give any indication about the accuracy of the prediction, but is used as a measure of diversity in our predictions.
\paragraph{Datasets} 
We first evaluate on the standard datasets of 7-Scenes \cite{shotton2013scene} and Cambridge Landmarks \cite{kendall2015posenet} that have both been widely used to evaluate camera localization methods. Both datasets consist of RGB frames with associated ground truth camera poses and provide training as well as test sequences. In addition, we created synthetic as well as real datasets, that are specifically designed to contain repetitive structures and allow us to assess the real benefits of our approach in ambiguous environments. For synthetic data we render table models from 3DWarehouse\footnote{https://3dwarehouse.sketchup.com/} and create camera trajectories, such that ambiguous views are ensured to be included in our dataset. In particular we create a circular movement around the object. Specifically we use a \textit{dining table} and a \textit{round table} model with discrete modes of ambiguities. In addition, we create highly ambiguous real scenes using Google Tango and the graph-based SLAM approach RTAB-Map \cite{labbe2019rtab}. We acquire RGB and depth images as well as distinct ground truth camera trajectories for training and testing. We also obtain a reconstruction of those scenes. However, note that only the RGB images and corresponding camera poses are required to train our model and the reconstructions are used for visualization only. In particular, our training and test sets consist of 2414 and 1326 frames, respectively. Note that our network sees a single pose label per image.
\paragraph{Baselines and state of the art} 
We compare our approach to current direct camera pose regression methods, PoseNet \cite{kendall2017geometric} and MapNet \cite{brahmbhatt2018geometry}, that regress a single pose prediction with neural networks. More importantly, we assess our performance against two state-of-the-art approaches, namely BayesianPoseNet \cite{kendall2016modelling} and VidLoc \cite{clark2017vidloc}, that are most related to our work and predict a distribution over the pose space by using dropout and mixture density networks, respectively.
We further include the unimodal predictions of UBN, as well as BMMs, MBN-MB, trained using mixture density networks~\cite{bishop1994mixture,Gilitschenski2020Deep} as baseline models.
\begin{figure}[t]\footnotesize 
	\centering
	\begin{subfigure}[b]{0.485\textwidth}
		\includegraphics[width=\textwidth]{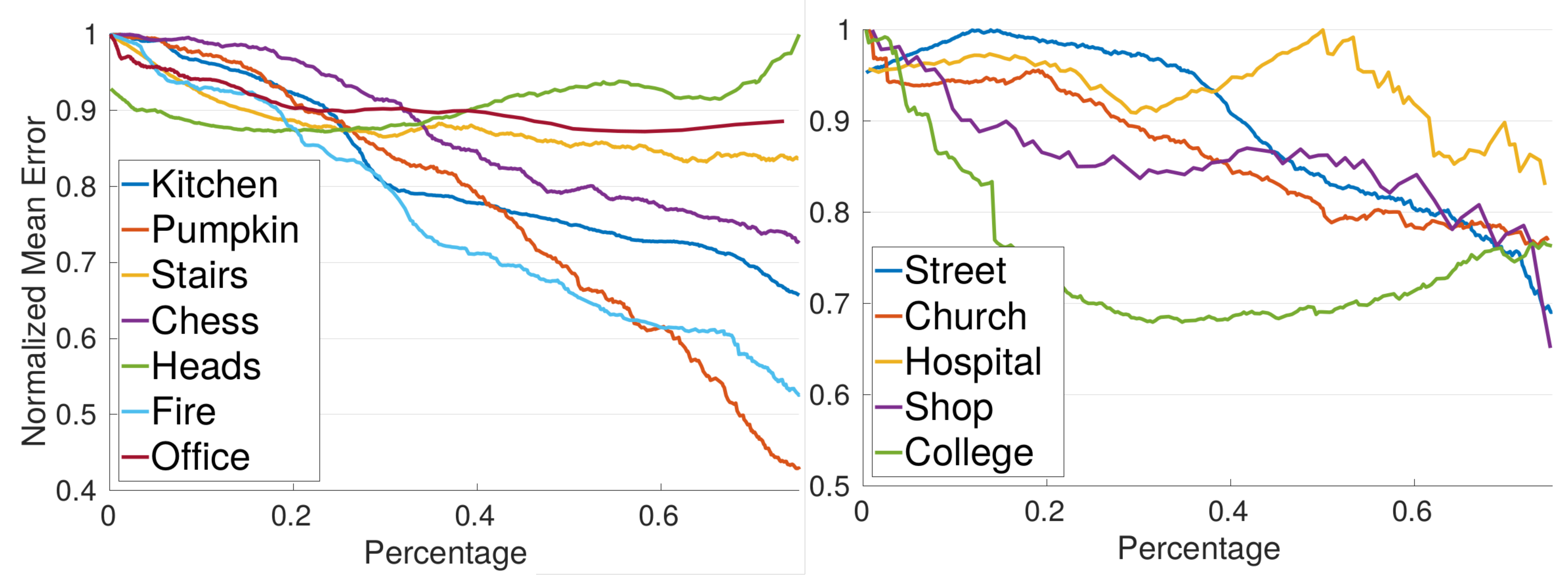}
		\caption{\footnotesize Rot. Uncertainty}
	\end{subfigure}\hfill
	\begin{subfigure}[b]{0.485\textwidth}
		\includegraphics[width=\textwidth]{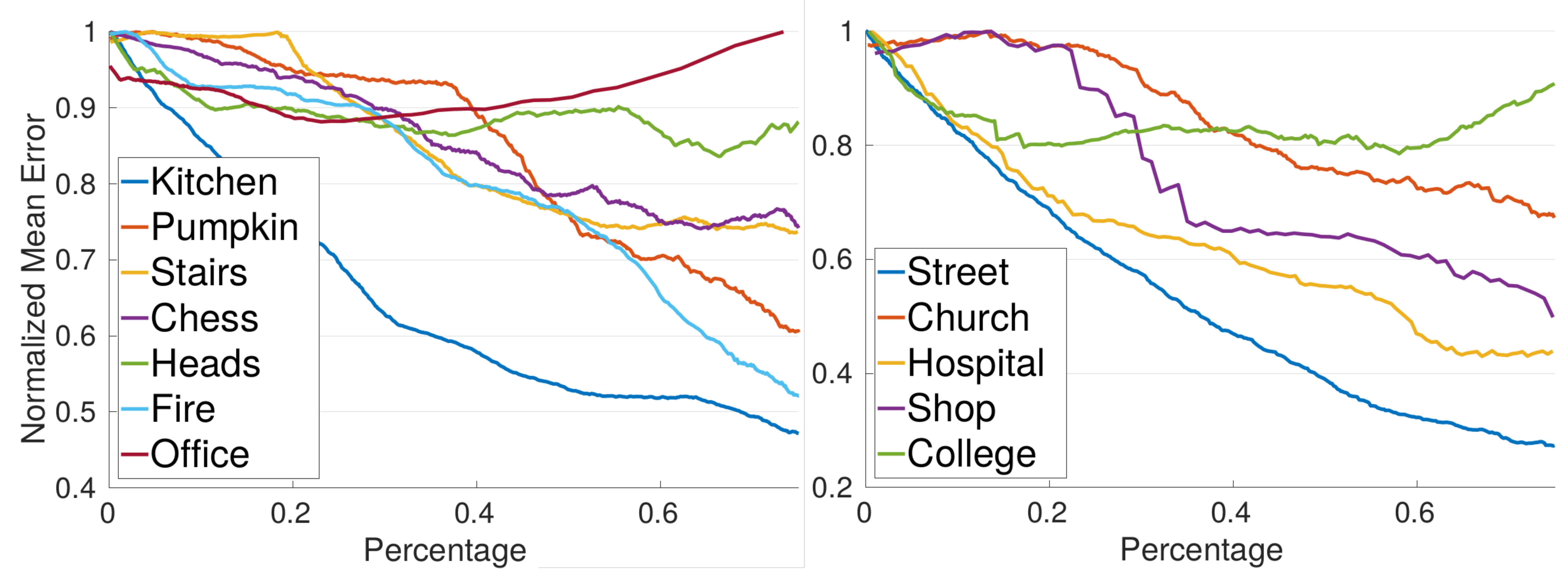}
		\caption{\footnotesize Trans. Uncertainty}
	\end{subfigure}
	\caption{Uncertainty evaluation on the 7-Scenes and Cambridge Landmarks datasets, showing the correlation between predicted uncertainty and pose error. Based on the entropy of our predicted distribution uncertain samples are gradually removed. We observe that as we remove the uncertain samples the overall error drops indicating a correlation between our predictions and the actual erroneous estimations on most scenes.}
	\label{fig:uncertainty_error}
\end{figure}

\subsection{Evaluation in non-ambiguous scenes}
We first evaluate our method on the publicly available 7-Scenes \cite{shotton2013scene} and Cambridge Landmarks \cite{kendall2015posenet} datasets. As most of the scenes contained in these datasets do not show highly ambiguous environments, we consider them to be non-ambiguous. Though, we can not guarantee that some ambiguous views might arise in these datasets as well, such as in the \textit{Stairs} scene of the 7-Scenes dataset. Both datasets have extensively been used to evaluate camera pose estimation methods. Following the state of the art, we report the median rotation and translation errors, the results of which can be found in~\cref{table:results}. In comparison to methods that output a single pose prediction (\eg~PoseNet \cite{kendall2017geometric} and MapNet \cite{brahmbhatt2018geometry}), our methods achieves similar results. Yet, our network provides an additional piece of information that is the uncertainty. On the other hand, especially in translation our method outperforms uncertainty methods, namely BayesianPoseNet \cite{kendall2016modelling} and VidLoc \cite{clark2017vidloc}, on most scenes.

\paragraph{Uncertainty evaluation}
One benefit of our method is that we can use the resulting variance of the predicted distribution as a measure of uncertainty in our predictions. The resulting correlation between pose error and uncertainty can be seen in~\cref{fig:uncertainty_error}, where we gradually remove the most uncertain predictions and plot the mean error for the remaining samples. The strong inverse correlation between the actual errors vs our confidence on most scenes shows that whenever our algorithm labels a prediction as uncertain it is also likely to be a bad estimate.

It has been shown that current direct camera pose regression methods still have difficulties in generalizing to views that differ significantly from the camera trajectories seen during training \cite{sattler2019understanding}. However, as we will show in the experiments section of our paper, these methods in addition suffer from ambiguities arising in the scene. Therefore, we analyze the performance of direct regression methods in a highly ambiguous environment. In this scenario even similar trajectories can confuse the network and easily lead to wrong predictions, for which our method proposes a solution.

\subsection{Evaluation in ambiguous scenes}
We start with quantitative evaluations on our synthetic as well as real scenes before showing qualitative results of our and the baseline methods. As recent results suggest that the ResNet network architecture is more effective in direct camera pose regression methods \cite{brahmbhatt2018geometry}, we exchange the initially used GoogleNet architecture with a ResNet for the state-of-the-art methods. In the following, we thus refer to BayesianPoseNet as MC-Dropout.

\begin{table*}[htbp]
	\begin{center}
		\caption{ Ratio of correct poses on our ambiguous scenes for several thresholds. We report the results of our MBN as MBN-$M$, where $M$ is the number of hypotheses used.}
		\resizebox{\textwidth}{!}{		
			\begin{tabular}{lccc|ccccccc}
			    \toprule
				& Threshold & \vtop{\hbox{\strut PoseNet}\hbox{\strut ~~~\cite{kendall2015posenet}}} & \vtop{\hbox{\strut MC-Dropout }\hbox{\strut ~~~~~~~\cite{kendall2016modelling}}} & UBN & MBN-MB
			    & MBN-CE & MBN-5 & MBN-10 & MBN-25 & MBN-50 \\
				\noalign{\smallskip}
				\midrule
				\noalign{\smallskip}
				& 10$^\circ$ / 0.1m & 0.19 & 0.39 & 0.29 & 0.24 & 0.35 & 0.40 & \textbf{0.48} & 0.35 & 0.39\\
				Blue Chairs (A) &15$^\circ$ / 0.2m & 0.69 & 0.78 & 0.73 & 0.75 & 0.81 & 0.85 & \textbf{0.92} & 0.80 & 0.79\\
				&20$^\circ$ / 0.3m & 0.90 & 0.88 & 0.86 & 0.80 & 0.82 & 0.89 & \textbf{0.96} & 0.87 & 0.85\\
				\midrule
				&10$^\circ$ / 0.1m & 0.0 & 0.04 & 0.02 & 0.01 & 0.05 & 0.03 & 0.07 & \textbf{0.08} & 0.03\\
				Meeting Table (B) &15$^\circ$ / 0.2m & 0.05 & 0.13 & 0.12 & 0.07 & 0.28 & 0.26 & \textbf{0.33} & 0.31 & 0.32\\
				&20$^\circ$ / 0.3m & 0.10 & 0.22 & 0.19 & 0.10 & 0.39 & 0.34 & \textbf{0.42} & 0.38 & 0.41\\
				\midrule
				&10$^\circ$ / 0.1m & 0.14 & 0.13 & 0.11 & 0.04 & 0.18 & \textbf{0.20} & 0.18 & 0.18 & 0.17 \\
				Staircase (C) &15$^\circ$ / 0.2m & 0.45 & 0.32 & 0.48 & 0.15 & \textbf{0.50} & 0.47 & 0.47 & 0.47 & 0.49 \\
				&20$^\circ$ / 0.3m & 0.60 & 0.49 & 0.62 & 0.25 & \textbf{0.68} & 0.64 & 0.66 & 0.64 & 0.64\\
				\midrule
				&10$^\circ$ / 0.1m & 0.07 & 0.02 & 0.06 & 0.06 & 0.09 &0.10 & 0.06 & 0.09 & \textbf{0.11} \\
				Staircase Extended (D) &15$^\circ$ / 0.2m & 0.31 & 0.14 & 0.26 & 0.21 & 0.39 & 0.43 & 0.38 & 0.44 & \textbf{0.46}\\
				&20$^\circ$ / 0.3m & 0.49 & 0.31 & 0.41 & 0.32 & 0.58 & 0.59 & 0.60 & 0.62 & \textbf{0.64}\\
				\midrule
				&10$^\circ$ / 0.1m & 0.37 & 0.18 & 0.11 & 0.06 & 0.35 & \textbf{0.39} & 0.38 & 0.35 & 0.37\\
				Seminar Room (E)&15$^\circ$ / 0.2m & 0.81 & 0.58 & 0.36 & 0.23 & 0.83 & 0.77 & 0.78 & \textbf{0.84} & 0.80\\
				&20$^\circ$ / 0.3m & 0.90 & 0.78 & 0.57 & 0.40 & \textbf{0.95} & 0.88 & 0.93 & 0.94 & 0.92\\
				\midrule\midrule
				&10$^\circ$ / 0.1m & 0.15 & 0.15 & 0.12 & 0.08 & 0.20 & 0.23 & \textbf{0.24} & 0.21 & 0.22\\
				Average &15$^\circ$ / 0.2m & 0.46 & 0.39 & 0.39 & 0.28 & 0.56 & 0.56 & \textbf{0.58} & 0.57 & 0.57\\
				&20$^\circ$ / 0.3m & 0.60 & 0.54 & 0.53 & 0.37 & 0.68 & 0.67 & \textbf{0.71} & 0.69 & 0.69\\
				\bottomrule
			\end{tabular}
		}
		\label{table:ambigious}
	\end{center}
\end{table*}

\begin{table}[t]
    \centering

    \caption{SEMD of our method and MC-Dropout indicating highly diverse predictions by our method in comparison to the baseline. Capital letter refer to the scenes of our ambiguous relocalization dataset with A: Blue Chairs, B: Meeting Table, C: Staircase, D: Staircase Extended and E: Seminar Room.}
    \begin{tabular}{lccccc}
    \toprule
			Method/Scene & A & B & C & D & E \\
			\midrule
			MC-Dropout & 0.06 & 0.11 & 0.13 & 0.26 & 0.10 \\
			MBN-CE & 1.19 & 2.13 & 2.04 & 3.81 & 1.70 \\
			MBN & \textbf{1.20} & \textbf{2.53} & \textbf{2.24} & \textbf{4.35} & \textbf{2.22} \\
			\bottomrule
	\end{tabular}

	\label{tab:semd}
\end{table}

\subsubsection{Explicit mixture coefficient learning}
We first evaluate explicit learning of the mixture coefficients and present results of MBN-CE, earlier introduced in \cite{bui2020eccv}, before providing additional evaluations of MBN, as proposed in this paper.
\paragraph{Quantitative evaluations}
We specifically created our synthetic dataset to contain a discrete set of modes such that we can easily identify ambiguous views. In particular, we know that there are two and four possible modes for each image in the \textit{dining} and \textit{round} table scenes respectively. Hence, to analyze the predictions of our model and its ability to avoid mode collapse we compute the accuracy of correctly detected modes of the true posterior. A mode is considered as found if there exists one pose hypothesis that falls into a certain rotational (5$^\circ$) and translational (10\% of the diameter of ground truth camera trajectory) threshold of it. In the dining-table scene, we observe that MC-Dropout obtains an accuracy of 50\%, finding one mode for each image, whereas the accuracy of MBN-CE on average achieves 96\%. On the round-table scene, our model shows an average detection rate of $99.1\%$, in comparison to $24.8\%$ of MC-Dropout, which suffers from mode collapse and even though it is able to predict multiple hypotheses, they tend to reside around one particular mode.
\begin{figure*}[th!]
	\centering
	\includegraphics[width=\textwidth]{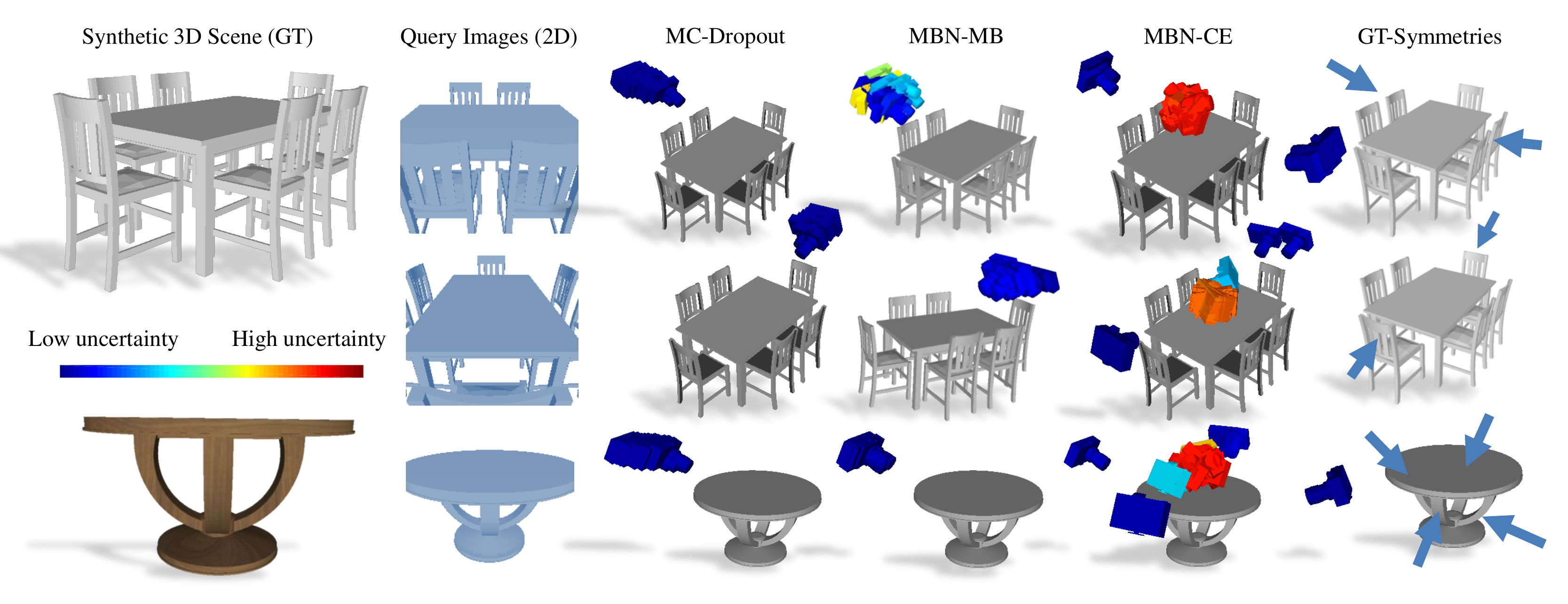}
	\caption{Qualitative results on our synthetic \textit{dining} and \textit{round table} datasets. Camera poses are colored by uncertainty. %\cSlo{Put somewhere the bar indicating how %uncertainty is color coded. Blue very confidetn, red not %confident.} 
	Viewpoints are adjusted for best perception.}
	\label{fig:synth}
\end{figure*}
\begin{figure*}[h!]
	\centering
	\includegraphics[width=\textwidth]{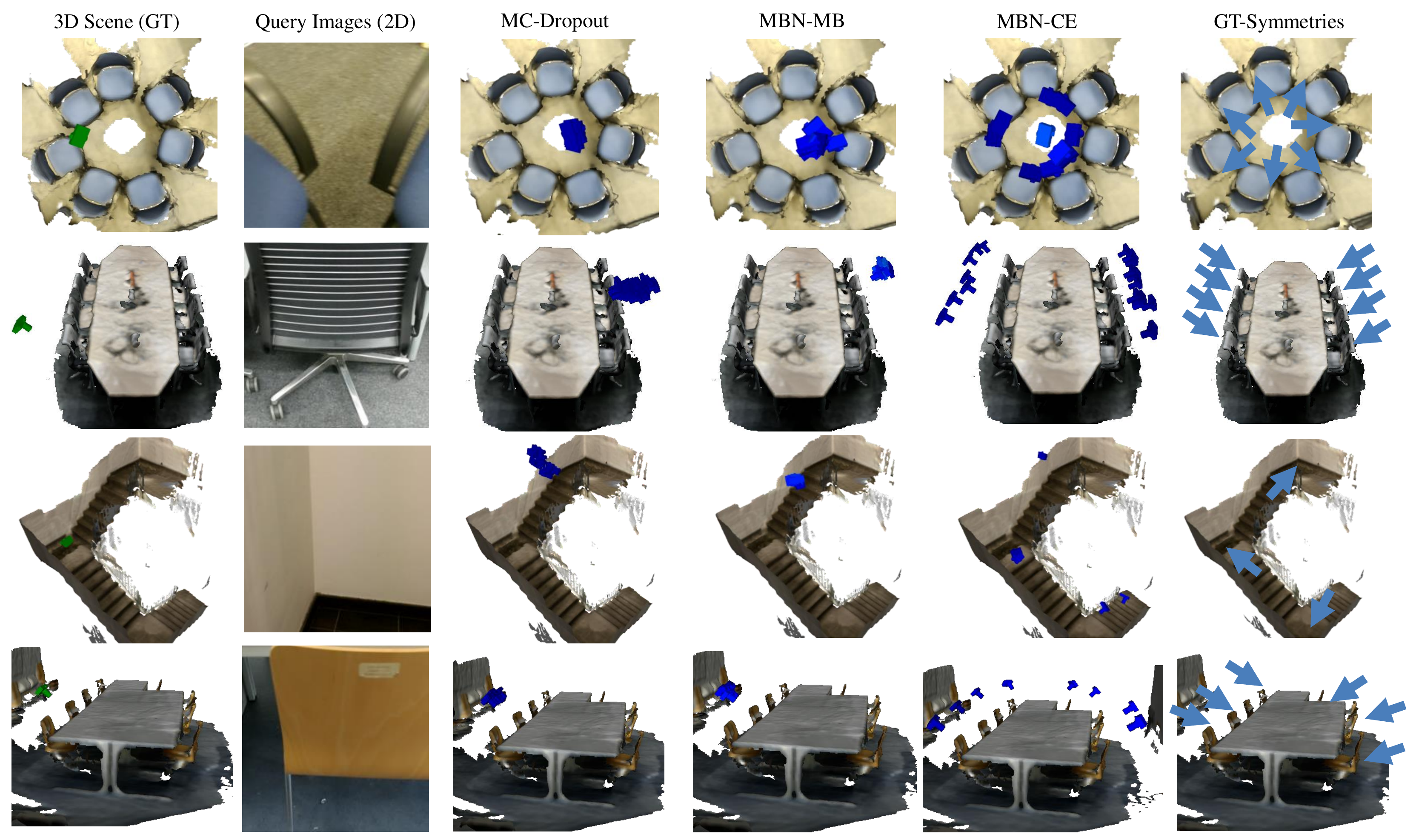}
	\caption{Qualitative results in our ambiguous relocalization dataset. For better visualization, camera poses have been pruned by their uncertainty values.}
	\label{fig:real}
\end{figure*}
\begin{figure*}[t]
	\centering
	\includegraphics[width=\textwidth]{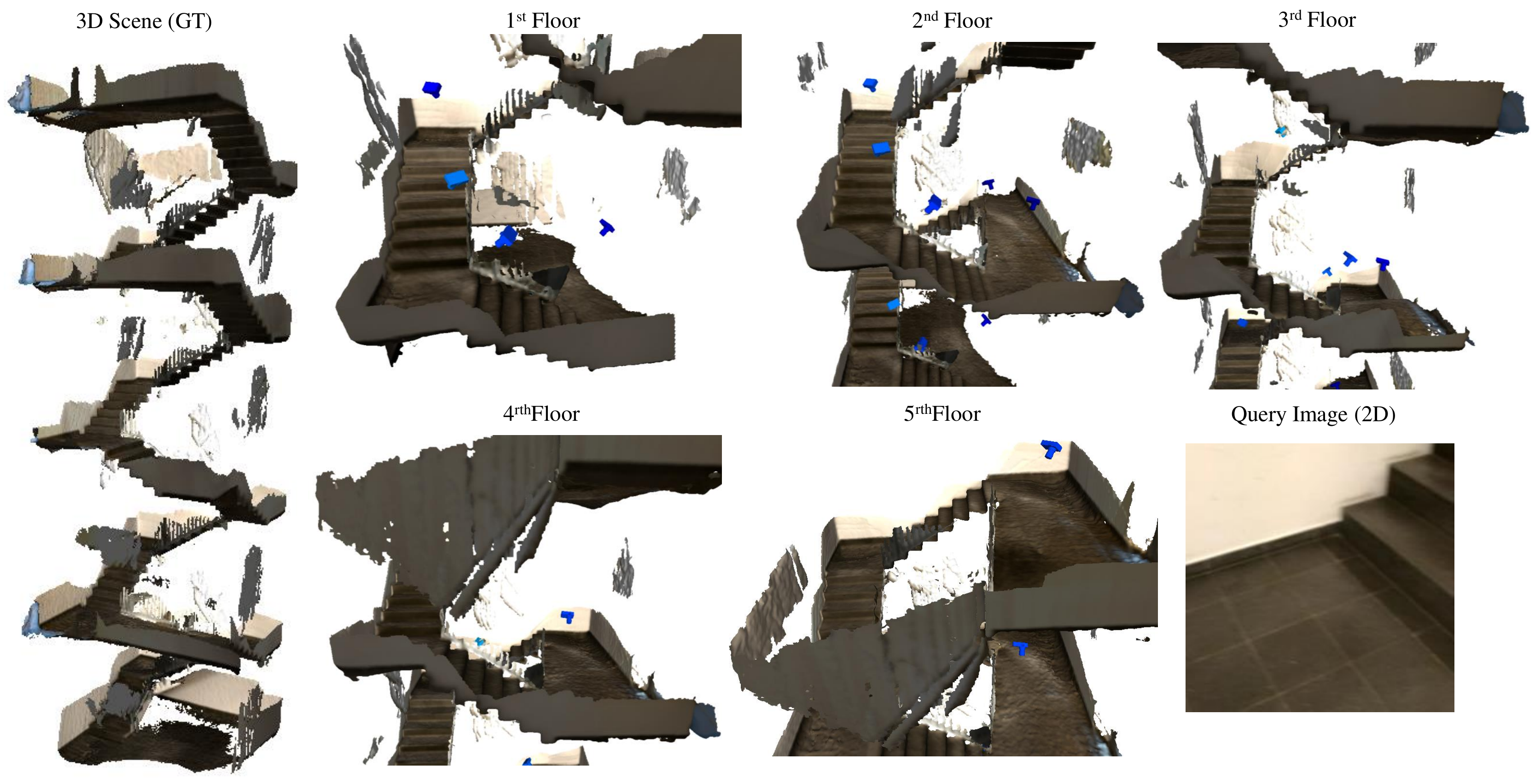}
	\caption{Qualitative results of our model on the \textit{Staircase Extended} scene. We show a reconstruction of the scene as well as predicted camera poses for a given query image.}
	\label{fig:real_stairs}
\end{figure*}
\begin{table}[t]

  \centering
  \caption{ Ratio of correctly detected modes for various translational thresholds (in meters). A refers to the \textit{Blue Chairs} scene, whereas B stands for the \textit{Meeting Table} scene.}
    \begin{tabular}{c|l|cccc}
    \toprule
    \multicolumn{1}{c}{Scene} & \multicolumn{1}{c}{Method} & 0.1 & 0.2 & 0.3 & 0.4 \\
    \midrule
    \multirow{2}[2]{*}{A} & MC-Dropout & 0.11  & 0.15  & 0.16  & 0.16 \\
          & MBN-CE & 0.36 & \textbf{0.79} & \textbf{0.80} & \textbf{0.80} \\
          & MBN & \textbf{0.42} & 0.76 & 0.77 & 0.77 \\
    \midrule
    \multirow{2}[1]{*}{B} & MC-Dropout & 0.04  & 0.07  & 0.09  & 0.11 \\
          & MBN-CE & 0.10 & \textbf{0.43} & \textbf{0.63} & \textbf{0.73} \\
          & MBN & \textbf{0.12} & 0.41 & 0.60 & \textbf{0.73} \\

          \bottomrule

    \end{tabular}%
  \label{table:mode_detection}

\end{table}

On our real scenes, we report the recall, where a pose is considered to be correct if both the rotation and translation errors are below a pre-defined threshold.
Thanks to the diverse mode predictions of our MBN-CE, which is indicated by the high SEMD shown in~\cref{tab:semd}, we are able to improve upon our baselines predictions.
Further, by a semi-automatic labeling procedure, we are able to extract ground truth modes for the \textit{Blue Chairs} and \textit{Meeting Table} scenes. For that aim, we train an autoencoder on reconstructing the input images and use the resulting feature descriptors to obtain the nearest neighbor camera poses. Then, we cluster the resulting camera poses using a Riemannian Mean Shift algorithm~\cite{subbarao2009nonlinear} and use the centroids of the resulting clusters as "ground truth" modes. We visually verify the results. This way, we can evaluate the entire set of predictions against the ground truth.~\cref{table:mode_detection} shows the percentage of correctly detected modes for our method in comparison to MC-Dropout when evaluating with these ground truth modes. The results again support our findings, that MC-Dropout suffers from mode collapse, such that even with increasing threshold the number of detected modes does not increase significantly.

\paragraph{Qualitative evaluations}
Qualitative results of our proposed model on our synthetic table datasets are shown in~\cref{fig:synth}. MC-Dropout as well as our finite mixture model, \textit{MBN-MB}, suffer from mode collapse. In comparison, the proposed MHP model is able to capture plausible, but diverse modes as well as associated uncertainties. In contrast to other methods that obtain an uncertainty value for one prediction, we obtain uncertainty values for each hypothesis. This way, we could easily remove non-meaningful predictions, that for example can arise in the WTA and RWTA training schemes.

\cref{fig:real} shows qualitative results on our ambiguous real scenes. Again, MC-Dropout and MBN-MB suffer from mode collapse. Moreover, these methods are unable to predict reasonable poses given highly ambiguous query images. That is most profound in our \textit{Meeting Table} scene, where due to its symmetric structure the predicted camera poses fall on the opposite side of the ground truth one. 

\subsubsection{Implicit mixture coefficients learning}
We now evaluate our extended method, MBN, that allows for implicit learning of the mixture coefficients without succumbing to the pitfalls of ordinary mixture density networks such as mode collapse. \cref{table:ambigious} shows the accuracy of our baseline methods in comparison to ours for various thresholds. Especially on our \textit{Meeting Table} scene, it can be seen that the performance of direct camera pose regression methods that suffer from mode collapse drops significantly due to the presence of ambiguities in the scene. In addition, we are able to improve upon our baseline model, MBN-CE, in overall performance, as well as in SEMD as reported in \cref{tab:semd}. Further, in comparison to MC-Dropout our model is able to provide diverse predictions and capture multiple modes, which is indicated by the high Oracle accuracy, see \cref{table:oracle}.

\begin{table}[t]
	\begin{center}
		\caption{ Average ratio of correct oracle poses on our ambiguous relocalization scenes for several thresholds and numbers of predicted pose hypothesis $M$, indicated as Oracle-$M$.}
		\resizebox{0.5\textwidth}{!}{		
			\begin{tabular}{cccccc}
			    \toprule
				Threshold & \vtop{\hbox{\strut MC-Dropout }\hbox{\strut ~~~~Oracle}} & \vtop{\hbox{\strut MBN-}\hbox{\strut Oracle-5}} & \vtop{\hbox{\strut MBN- }\hbox{\strut Oracle-10}}& \vtop{\hbox{\strut MBN-}\hbox{\strut Oracle-25}}& \vtop{\hbox{\strut MBN- }\hbox{\strut Oracle-50}}\\
				\noalign{\smallskip}
				\midrule
				\noalign{\smallskip}
				10$^\circ$ / 0.1m & \textbf{0.28} & 0.26 & 0.27 & 0.27 &  0.26\\
				15$^\circ$ / 0.2m &  0.60 & 0.56 & 0.60 & \textbf{0.67} & \textbf{0.67}\\
				20$^\circ$ / 0.3m &  0.70 & 0.69 & 0.75 & 0.80 & \textbf{0.84}\\
				\bottomrule
			\end{tabular}
		}
		\label{table:oracle}
	\end{center}
\end{table}

\begin{table*}[t]
	\begin{center}

		\caption{Averaged ratio of correct poses for different backbone networks over all scenes of our ambiguous relocalization dataset.}

		\setlength{\tabcolsep}{10pt}
			\begin{tabular}{lccccccc}
			     & Threshold & PoseNet & UBN & MBN-MB & MC-Dropout & MBN-CE & MBN \\
				\noalign{\smallskip}
				\midrule
				& 10$^\circ$ / 0.1m & 0.15 & 0.12 & 0.08 & 0.15 & 0.20 & \textbf{0.22}\\
				ResNet-34 & 15$^\circ$ / 0.2m & 0.46 & 0.39 & 0.28 & 0.39 & 0.56 & \textbf{0.57}\\
				& 20$^\circ$ / 0.3m & 0.60 & 0.53 & 0.37 & 0.54 & 0.68 & \textbf{0.69}\\
				\midrule
				 & 10$^\circ$ / 0.1m& 0.15 & 0.16 & 0.09 & 0.15 & 0.19 & \textbf{0.20}\\
				ResNet-18 & 15$^\circ$ / 0.2m & 0.47 & 0.42 & 0.29 & 0.39 & 0.52 & \textbf{0.53}\\
				& 20$^\circ$ / 0.3m & 0.60 & 0.54 & 0.39 & 0.54 & \textbf{0.66} & 0.64\\
				\midrule
				& 10$^\circ$ / 0.1m & \textbf{0.20} & 0.15 & 0.10 & 0.15 & \textbf{0.20} & \textbf{0.20}\\
				ResNet-50 & 15$^\circ$ / 0.2m & 0.49 & 0.36 & 0.30 & 0.40 & \textbf{0.55}&0.52\\
				& 20$^\circ$ / 0.3m & 0.62 & 0.53 & 0.38 & 0.53 & \textbf{0.69} & 0.67\\
				\midrule
				& 10$^\circ$ / 0.1m & 0.11 & 0.10 & 0.11 & 0.08 & \textbf{0.18} & 0.17\\
				Inception-v3 & 15$^\circ$ / 0.2m & 0.38 & 0.33 & 0.38 & 0.31 & 0.49 & \textbf{0.50}\\
				& 20$^\circ$ / 0.3m & 0.55 & 0.53 & 0.52 & 0.49 & \textbf{0.63} &\textbf{ 0.63}\\
				\noalign{\smallskip}
				\bottomrule
			\end{tabular}

		\label{table:backbone}
	\end{center}
\end{table*}

\paragraph{Backbone network} To evaluate the effect of different network architectures on our model, we change the backbone network of ours and the state-of-the-art baseline methods. As most of the recent state-of-the-art image based localization methods \cite{balntas2018relocnet, brahmbhatt2018geometry, peretroukhin2019probabilistic} use a version of ResNet, we compare between ResNet variants: ResNet-18, ResNet-34 and ResNet-50 and Inception-v3~\cite{szegedy2016rethinking}. All networks are initialized from an ImageNet~\cite{deng2009imagenet} pre-trained model. We report our findings in~\cref{table:backbone}. When comparing the performance of different ResNet variants all networks showed on average similar accuracy. Overall, naturally all methods are slightly dependant on the features that serve as input to the final pose regression layers. However, regardless of the backbone network used, MBN-CE and MBN show, on average, superior performance over the baseline methods.

\subsection{Choice of best branch}
The choice of the best branch in multiple hypothesis predictions depends on the chosen distance function comparing the prediction to the ground truth label.

\begin{figure}[t]
	\begin{center}
	    \includegraphics[width=0.4\textwidth]{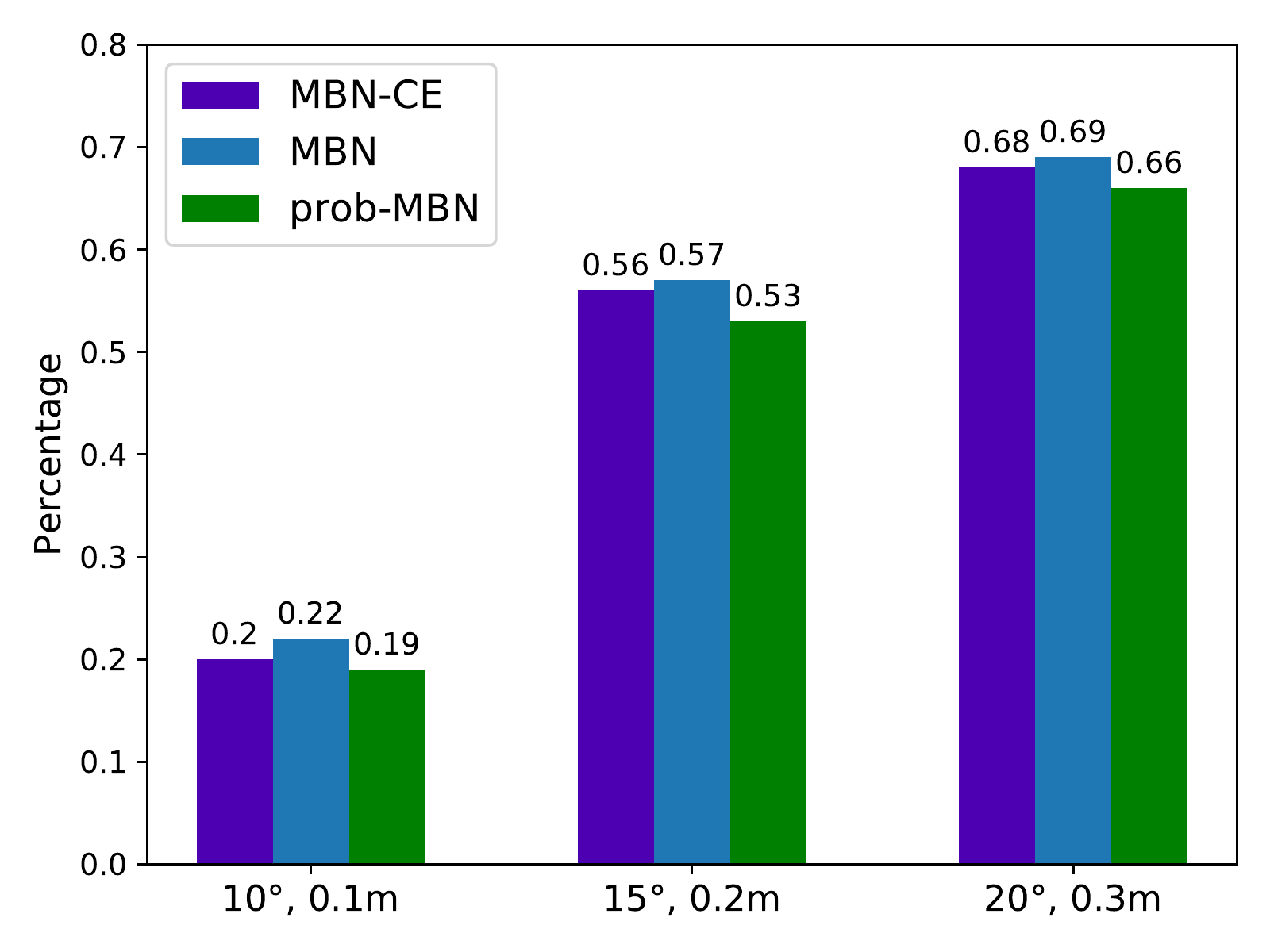}
		\caption{Influence on the choice of the best hypothesis on our MHP training scheme. We compare between MBN-CE and MBN ($l_1$ loss) and prob-MBN, i.e. choosing the best branch according to its probability (see \cref{eq:select_prob}).}
		\label{table:dist}
	\end{center}%\vspace{-7mm}
\end{figure}

In this work, we compare between the $l_1$ norm (see \cref{eq:select_l1}) and choosing the branch with highest probability density, as described in \cref{eq:select_prob}, and report the results in \cref{table:dist}. For our specific application, camera re-localization, we have found $l_1$ to outperform probability based choices.

\section{Application to Point Cloud Pose Estimation}

We continue to apply our \modelnames~on the task of class-level point cloud pose estimation. We assume that the class of the test object is known and an individual network is trained for each class. The pose of a point cloud can be expressed by a combination of rotation $\mathbf{R} \in SO(3)$ and translation $\mathbf{t} \in \R^3$. For 3D objects, translation can be canceled out using the centroid or a common anchor, whereas the rotation has to be estimated, which is also the main source of ambiguities in this application. Similarly, the latter quantity is parameterized by a unit quaternion $\hat{\q}\in \mathbb{H}_1$.

\begin{table*}[htbp]
  \centering
  \caption{Point cloud pose estimation results on ModelNet10. The values are scaled by $10^2$.}
  \resizebox{1.0\textwidth}{!}{
    \begin{tabular}{ccccccc|ccccc|cc}
    \toprule
          & L1    & Cosine & Ploss & PointNetLK & IT-Net & MC-Dropout & UBN   & MBN-5 & MBN-10 & MBN-25 & MBN-50 & MBN-CE & MBN-MB \\
    \midrule
    Bathtub & 4.126 & 7.141 & 2.262 & 10.110 & 11.015 & 5.994 & 1.064 & 0.728 & 0.557 & \textbf{0.490} & 0.504 & 0.790 & 0.805 \\
    Bed   & 1.815 & 3.049 & 1.610 & 12.330 & 7.106 & 1.907 & 0.918 & 0.331 & \textbf{0.235} & 0.267 & 0.332 & 0.790 & 0.656 \\
    Chair & 0.653 & 1.108 & 0.859 & 8.280 & 3.272 & 0.866 & 0.999 & 0.734 & \textbf{0.600} & 0.727 & 0.663 & 0.790 & 0.970 \\
    Desk  & 6.101 & 8.190 & 4.529 & 10.730 & 11.299 & 6.068 & 3.190 & 2.129 & \textbf{1.953} & 2.615 & 2.656 & 2.620 & 2.510 \\
    Dresser & 4.524 & 6.753 & 2.888 & 7.260 & 8.500 & 5.124 & 2.285 & 1.423 & \textbf{1.372} & 1.669 & 1.771 & 2.620 & 2.166 \\
    Monitor & 3.005 & 5.547 & 2.172 & 13.230 & 11.184 & 2.980 & 2.038 & 1.009 & 0.988 & \textbf{0.984} & 1.169 & 1.150 & 1.243 \\
    Night Stand & 3.661 & 4.144 & 2.987 & 5.700 & 6.348 & 3.535 & 1.943 & 1.602 & 1.282 & \textbf{1.248} & 1.281 & 1.600 & 2.066 \\
    Sofa  & 0.727 & 1.368 & 0.786 & 12.460 & 3.763 & 0.820 & 0.620 & \textbf{0.314} & 0.327 & 0.352 & 0.408 & 0.300 & 0.444 \\
    Table & 10.825 & 16.820 & 1.253 & 16.550 & 15.537 & 7.063 & 0.871 & \textbf{0.428} & 0.519 & 0.506 & 0.566 & 0.780 & 0.740 \\
    Toilet & 0.609 & 1.389 & 0.582 & 7.430 & 3.746 & 0.730 & 0.846 & 0.576 & 0.544 & 0.401 & \textbf{0.377} & 0.570 & 0.769 \\
    \midrule
    Average & 3.605 & 5.551 & 1.993 & 10.410 & 8.180 & 3.509 & 1.477 & 0.927 & \textbf{0.838} & 0.926 & 0.973 & 1.201 & 1.237 \\
    \bottomrule
    \end{tabular}%
  }%
  \label{tab:benchmark}%
\end{table*}%

\subsection{Implementation and Training Details}
Our \modelnames~and losses are implemented using Pytorch~\cite{pytorch}
and we use PointNet~\cite{qi2017pointnet} as the backend to process point
clouds. We train each class for 500 epochs and in each epoch 100
quaternions are randomly sampled to rotate the training objects. We set the
learning rate at 0.001 and use Adam solver~\cite{adam} to optimize the
network parameters.

\textit{Birdal Strategy} is used as the default way of
constructing $\V$. The default training loss for MBN is a combination of
\textit{Mixture Bingham Loss} and \textit{RWTA Loss} and the best component
for computing \textit{RWTA Loss} is chosen by probability.

\subsection{Experiment Setup}

\paragraph{Baselines} We extensively studied the state-of-the-art methods on pose/rotation estimation, however, due to the differences in the specific targeted applications, such as camera localization \cite{kendall2017geometric}, point cloud registration \cite{wang2019deep} or object pose estimation \cite{manhardt2019explaining}, and the accordingly varying network selections, it is very difficult to obtain a fair direct comparison. Therefore, we evaluated the loss functions commonly employed by the state of the art:
\begin{itemize}[noitemsep,leftmargin=*]
    \item \textbf{L1 loss}. The most popular one is $L_p\,(p=1,2)$ loss, and it has been most widely used in recent work for rotation regression~\cite{wang2019deep, wang2019prnet, aoki2019pointnetlk, kendall2017geometric, liao2019spherical}. Previous work has found that $L_1$ outperforms $L_2$ in general~\cite{kendall2017geometric}, so we mainly take $L_1$ loss for comparison: $\mathcal{L}_1 = ||\q - \hat{\q}||_1$.
    \item \textbf{Cosine loss}. A metric often used for measuring distance on the spherical manifold, cosine loss respects the vectorial nature of quaternions~\cite{mahendran20173d}: $\mathcal{L}_{\text{cos}} = 1 - |\q \cdot \hat{\q}|$.
    \item \textbf{PLoss}. Defined on the points rather than the rotations themselves, PLoss~\cite{xiang2018posecnn, yuan2018iterative} yields point-wise Euclidean distances: $\mathcal{L}_{\text{PLoss}} = ||\q \circ \x - \hat{\q} \circ \x||_2$.
\end{itemize}
We unify the above loss functions under a common framework for a fair comparison with our Bingham losses. The same network, dataset, training schemes and tasks are used to compute different losses. Note, when training with those losses, only the quaternion parts of our \modelabbrv~are trained. It is also possible to incorporate our Bingham losses in their tasks to train their networks, simply by adding more outputs to the last layer without changing the overall architecture.

To further showcase the power of the proposed algorithm, we include some extra independent state-of-the-art baselines.
In particular, PointNetLK~\cite{aoki2019pointnetlk} and IT-Net~\cite{yuan2018iterative} are two state-of-the-art deep learning-based algorithms for iteratively estimating the relative poses between pairs of point clouds. For a fair comparison, we pair the canonical and rotated point clouds to compose the input and try to predict the relative rotation. This notion is identical to what our MBN tries to predict. The remaining configurations are kept the same as in the original work.

\paragraph{Metrics} Due to the potential ambiguities in the point clouds, it is inappropriate to use angular error to measure the qualities of the predictions with regard to the ground truths as different poses might align the point clouds equally well. Instead, we measure the \textit{Chamfer distance} (CD) between the point clouds rotated by the ground truth and predicted pose. In order to show how multiple modes are captured by our mixture Bingham networks, we also measure \textit{Self-EMD} (SEMD)~\cite{makansi2019overcoming}, the earth movers distance~\cite{rubner2000earth} of turning a multi-modal distribution into a unimodal one.

\paragraph{Dataset} We choose ModelNet10~\cite{wu20153d} as the benchmark to conduct our evaluations. This dataset contains 10 classes, and objects from each class possess unique geometries as well as different level of symmetries. This creates an ideal situation for validating our method in terms of uncertainty and ambiguity. We conduct class-level object pose estimation on this dataset, and follow the original train/test split.

\subsection{Pose Estimation Evaluation}

\insertimageC{1.0}{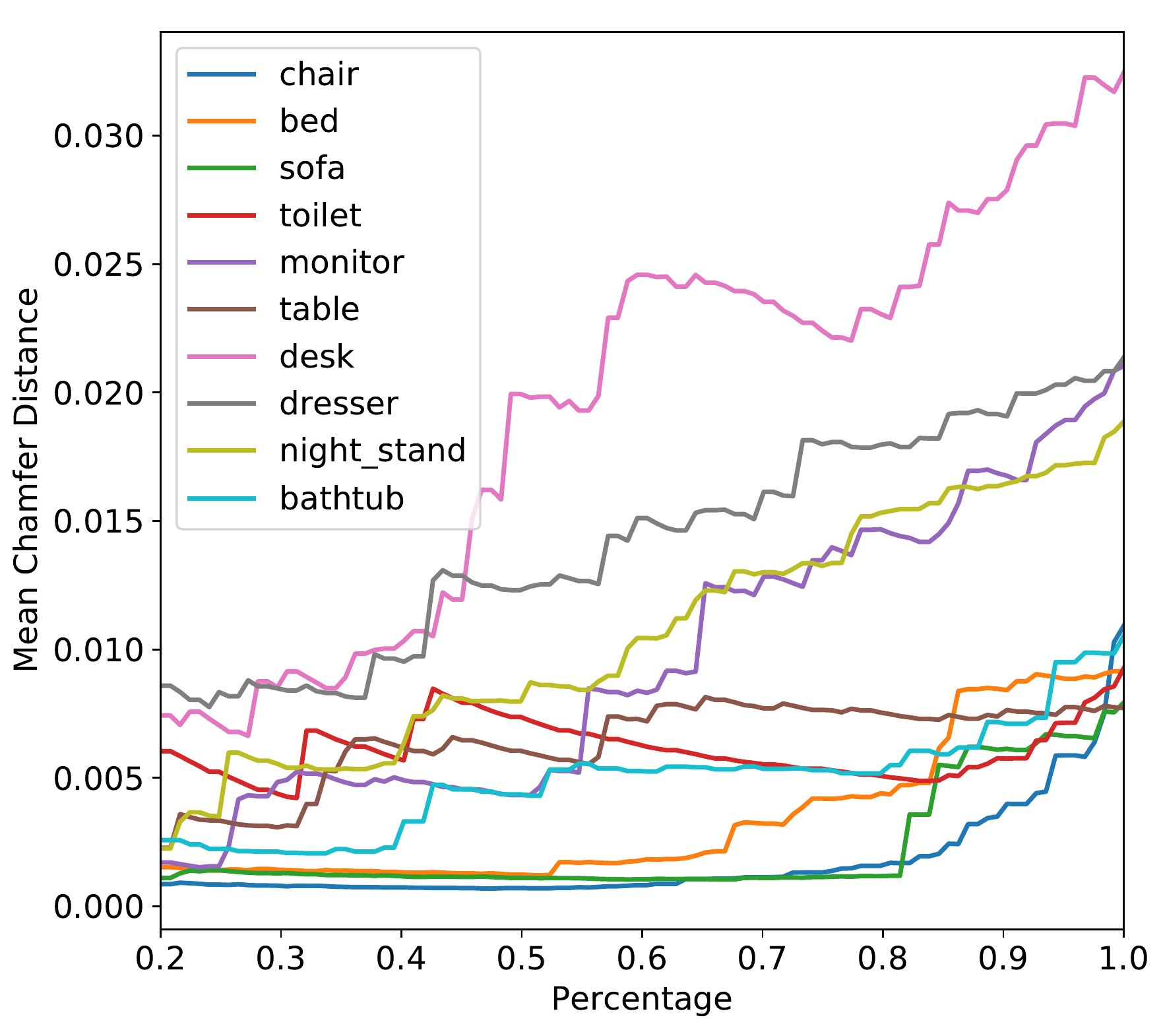}{As uncertainty threshold increases, the average CD of predictions whose uncertainties are below a threshold increases accordingly.}{fig:err_en}{t!}

UBN and MBN output continous distributions instead of discrete poses. To
evaluate their performance on the task of pose esimation, a single pose prediction
for UBN is decided by taking the mode of the predicted continuous Bingham
distribution, which equals the pose with highest chance in the according
probability space. Similarly, for MBN, the component with the largest weight
is selected and its mode serves as the single prediction in this evaluation.
\begin{figure}[t!]
    \centering
    \begin{subfigure}[b]{0.9\linewidth}
        \includegraphics[width=\textwidth]{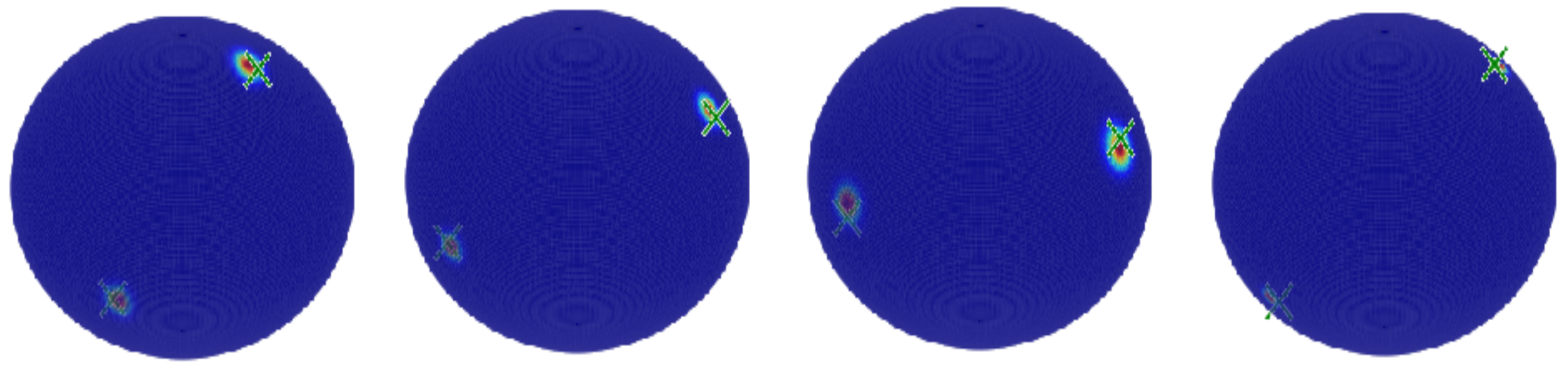}
        \caption{Good predictions tend to have small uncertainty.}
        \label{subfig:uncertainty_good}
    \end{subfigure}
    \begin{subfigure}[b]{0.9\linewidth}
        \includegraphics[width=\textwidth]{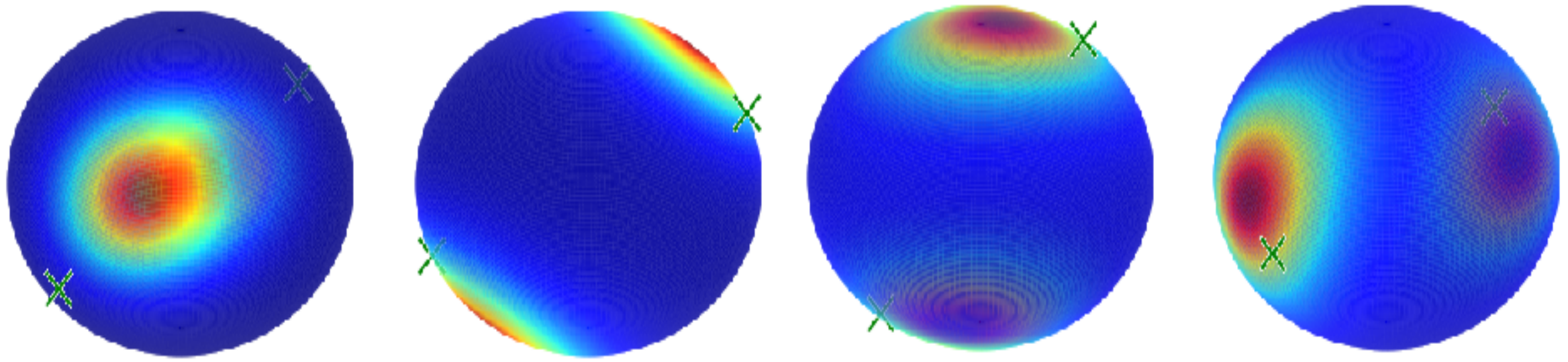}
        \caption{Bad predictions tend to have big uncertainty.}
        \label{subfig:uncertainty_bad}
    \end{subfigure}

    \caption{Bingham distributions generated by Unimodal Bingham Network. The ground truth poses are marked with ``$\times$''. Predictions of UBN are centered on the mode (the red-most region).}

    \label{fig:uncertainty}
\end{figure}
\paragraph{Quantitative evaluations}
~\Cref{tab:benchmark} showcases the average \textit{Chamfer Distance} results of all the baselines as well as our \modelname, including both UBN and MBN.

In comparison to all the baselines, our UBN not only achieves best performances across all the classes, but also provides an extra uncertainty information. We attribute the improvement to the fact that Bingham distribution enables an anisotropic distance taking into account the uncertainty in directions. This better reflects the real distance between the predictions and ground truth, thus lead to a better convergence.

However, like other baselines, UBN is incapable of dealing with ambiguities from point clouds with rotational symmetries.
This is powerfully demonstrated by the further improvements obtained by our
MBNs, where different modes triggering ambiguities
are captured by different components of the underlying multimodal Bingham
model.

PointNetLK~\cite{aoki2019pointnetlk} and IT-Net~\cite{yuan2018iterative} work in a similar way as the celebrated iterative closest point (ICP) algorithm~\cite{besl1992method} and they are proven to be robust methods. Yet as we can see from \cref{tab:benchmark}, they could not cope with the drastic changes in the poses that well and this leads to inferior performances on this evaluation. Different to all the baseline competitors, our MBN does not require a canonical point set to estimate the orientation of the input. Instead, it relies on a high-level abstracted/implicit canonical notion for the entire class. Moreover, this canonical form is learned from the training data, making the algorithm more robust and efficient. 

To see how number of components would impact the performance of MBNs, we
provide results for several versions of MBNs with 5, 10, 25 and 50 components
respectively. As the number of components increases from 5 to 10, a
further improvement can be clearly observed. However, the increase in the number of components (up to 25 and 50) does not translate to an increase in the performances. The existence of such sweet spot indicates the saturation of the mode diversity, \ie excessive over-parameterization of the modes causes complexity in learning. Nevertheless, overall, all our MBNs outperform the other baselines which lack the ability to handle ambiguity.

\begin{table}[t]
  \centering
  \caption{Filtering hypotheses for registration using uncertainty information. The first row is the threshold value below which the hypotheses survive. The second row is the registration recall. It reaches 77.7$\%$ if no hypotheses are filtered, resulting in the method of \cite{deng2019direct}. The third row is the percentage of filtered hypotheses. The lower the uncertainty threshold, the more hypotheses are dropped. With the aid of our uncertainty, more than 20$\%$ of the hypotheses could be neglected without harming the performance. Even when $54.1\%$ hypotheses are dropped, the performance decreases only by 4.7$\%$.}
    \begin{tabular}{cccccc}
    \toprule
    Threshold & 0.30  & 0.25  & 0.20  & 0.15  & 0.10 \\
    \midrule
    Ave. Recall & 0.777 & 0.773 & 0.769 & 0.742 & 0.730 \\
    Drop Rate & 0.184 & 0.221 & 0.280 & 0.388 & \textbf{0.541} \\
    \bottomrule
    \end{tabular}%
  \label{tab:registration}%
\end{table}%

\insertimageStar{0.9}{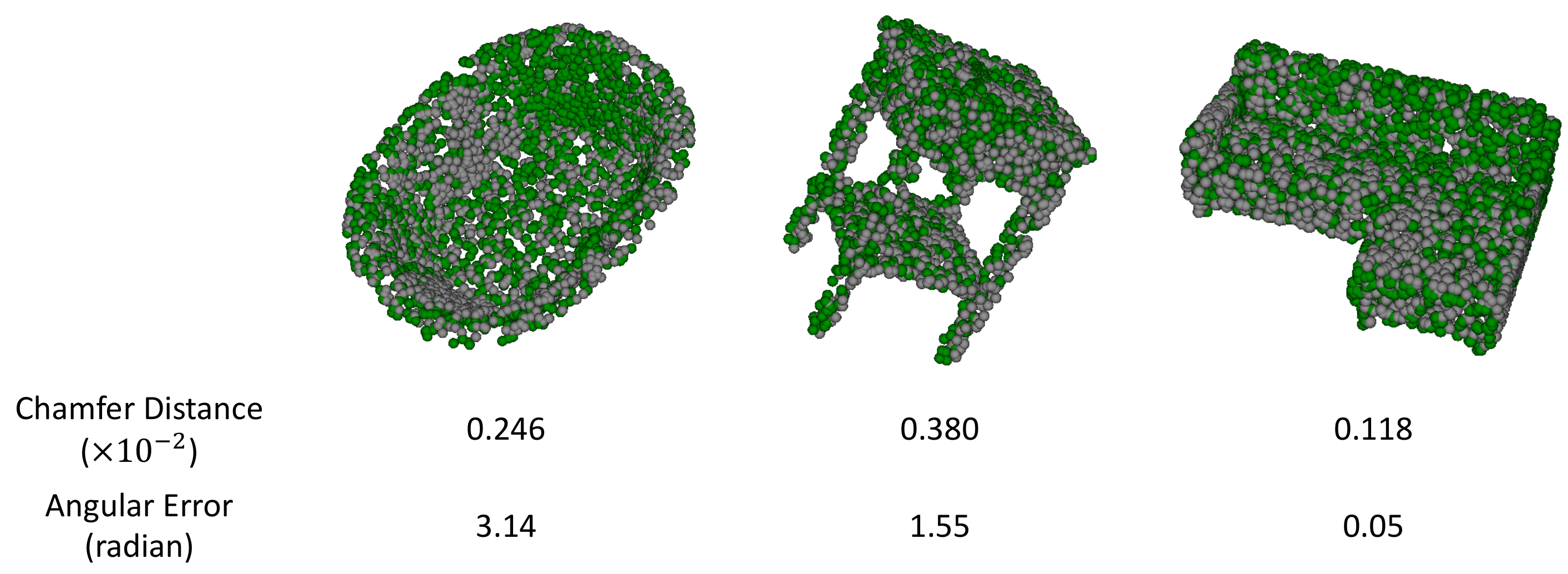}{The commonly used angular error would fail to be a good metric for the predictions which are different from the ground truth pose but still result in a good alignment for objects with symmetries. In this case, the chamfer distance better reflects the quality of the predictions. Gray points are from the ground truth point cloud and green ones from predicted point cloud.}{fig:cd_vs_ae}{t!}
\paragraph{Uncertainty Evaluation} To better understand how our uncertainty
gauge works, we plotted ~\cref{fig:err_en} by computing the average CDs of
predictions with uncertainties less than a varying number of
thresholds. As the uncertainty threshold decreases, the average CDs also
decrease accordingly, which indicates that predictions with less uncertainty
tend to align the point clouds better with the ground truth. We visualize the
predicted Bingham distributions along with the ground truths in ~\cref{fig:uncertainty}. In general, good predictions signal clustered and peaked distributions, while bad ones
tend to spread and result in higher uncertainties.

To further demonstrate that our uncertainty information could be useful in practical 3D applications, we take the Partial Scan Registration task from \cite{deng2019direct}. In their work, local patches from two adjacent partial scans are first matched by learned descriptors. Each pair of patches is fed into another regression network to predict the relative rigid transformation between them (rotation only, as the translation part can be computed later by the distances between centers of the local patches). All the local patches are ideally aligned under the same relative pose. In the last stage of their pipeline, a pool of pose hypotheses is generated and exhaustively evaluated. The single best hypothesis is in the end picked as the final pose prediction to register the two partial scans. 

For our purpose, we add extra output units to their network to predict \textit{concentration parameters} for $\mathbf{\Lambda}$ and then train it with our Bingham loss. With our extra uncertainty information, it is possible to filter out some of the generated hypotheses that now are associated with high uncertainties and as a result accelerate the registration, without harming the performance much, as shown in \cref{tab:registration}.

\insertimageStar{0.98}{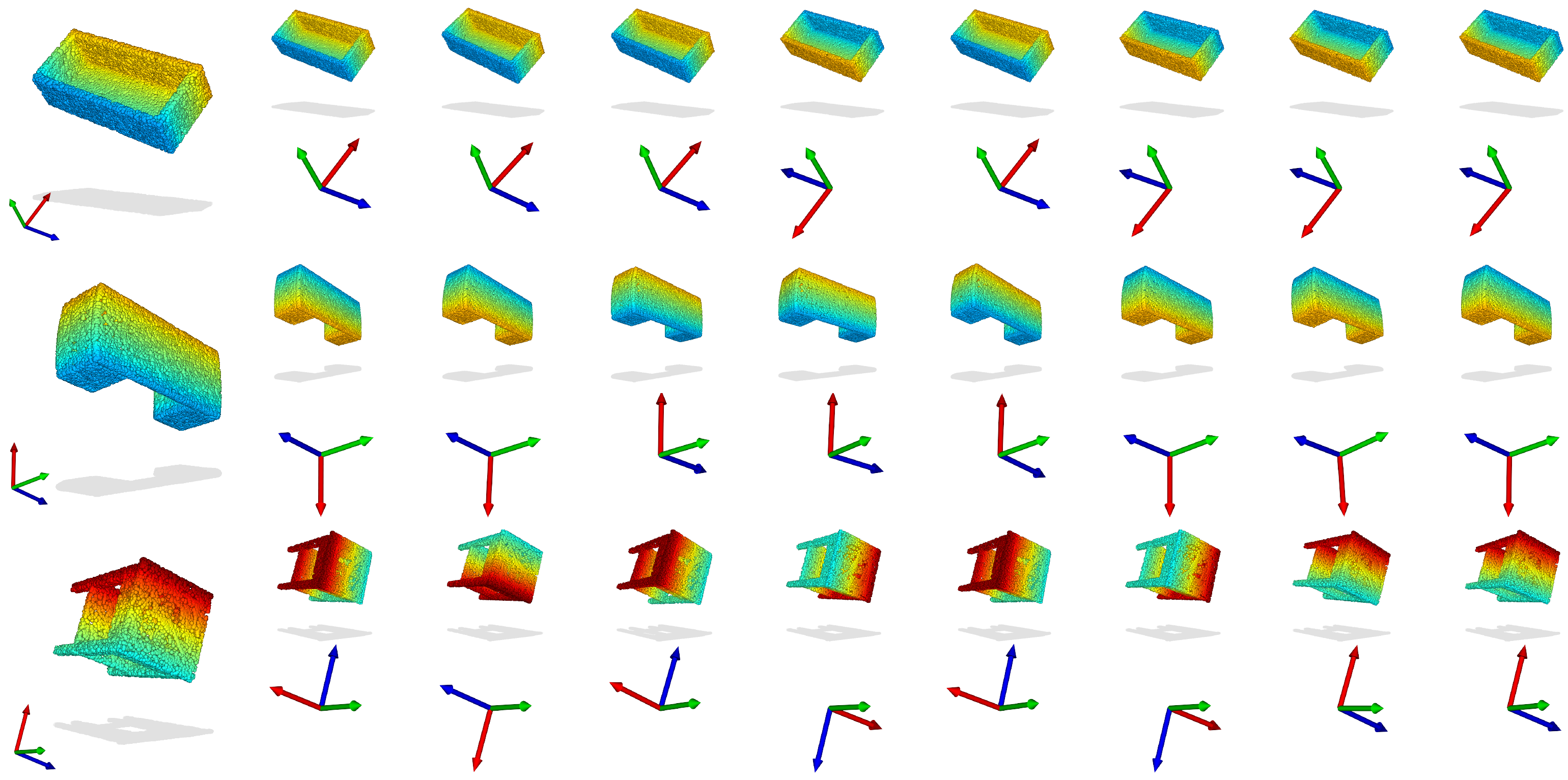}{Ambiguous poses could be well captured by different components of our Multimodal Bingham Network. The first column shows the object under ground truth poses. The rest of the columns show predicted poses (represented as local reference frames) and the correspondent rotated object. Point clouds are colored by the coordinates in the non-rotated version. }{fig:multi_pred}{t!}
\begin{table}[t]
    \centering
    \caption{Comparison between best branch selection criteria for MBN and the task of object pose estimation. Reported is the average Chamfer distance on the ModelNet10 dataset.}
    \begin{tabular}{ccc}
        \toprule
        Selection & L1    & Prob  \\
        \midrule
        CD        & 1.057 & 0.973 \\
        \bottomrule
    \end{tabular}%
    \label{tab:pc_selection}%
\end{table}%
\paragraph{Qualitative evaluations}
\cref{fig:cd_vs_ae} shows the overlaps of point clouds rotated by ground truth poses and predicted poses. In all the displayed cases, the predictions achieve good alignment with the ground truths, which can be well indicted by the small chamfer distances. However, if we use angular error as the metric to qualify the results, even though it could still obtain small values for unambiguous objects (\eg sofa), but it might disregard the first two predictions with big errors due to the ambiguities in the objects.   

\cref{fig:multi_pred} demonstrates how MBN is able to capture different modes in the data when ambiguities are presented. For objects with rotational symmetries, diffrent poses which could equivalently align them are captured by different components. It is possible to further derive the symmetic axis of those objects based on the set of various predictions.

In cases where an object does not carry ambiguity, we can see from
\cref{fig:multi_pred_no_ambi} that all the components tend to agree on the
predictions. It shows that extra components are not burdonsome for non-ambiguous objects. This kind of predictions could be further utilized as a sign to indicate whether the given point cloud is rotational symmetric or not.

\subsection{Choice of the best branch}
In this application, we use probability as the default metric in RWTA to select the best branch according to \cref{eq:select_prob}.
To validate this decision, we compare the two strategies as described in ~\cref{eq:select_l1} and ~\cref{eq:select_prob}. As we can see from ~\cref{tab:pc_selection}, performance-wise the two criteria are close to each other. However, as the probabilities are anyway needed for the final RWTA loss, it reduces the total amount of computation by using them for the purpose of branch selection as well.

\section{Ablation Studies}
\label{sec:ablation}

We now provide further ablation studies that have been conducted for both applications, relocalization and object pose estimation. First, we evaluate the effect of various methods of constructing $\V$ and winner-takes-all training schemes proposed in current literature. We then include another rotation parameterization in our framework that has been specifically proposed for training with neural networks. We summarize our findings in common tables where the two subtables refer to a) relocalization and b) object pose estimation.

\subsection{Variants of Constructing $\V$}
As described in section \ref{sec:V} we compare between three ways of incorporating orientation learning into neural networks, namely Gram-Schmidt (G), Cayley Transform (C) and the method of Birdal et. al ~\cite{birdal2018bayesian} (B). We report our findings in Table \ref{tab:ab_V}.
In comparison to Gram Schmidt orthonormalization the remaining methods only require four parameters to be estimated
instead of the 16 entries of the matrix $\V$. In the case of camera localization for our UBN as well as multimodal MBN-MB we found the Skew-Symmetric construction to outperform both Gram-Schmidt and the employed method of Birdal et al. However, overall we have found the method of Birdal and our MBN network to achieve the best performance. This is further validated for the task of object pose estimation where our MBN model clearly outperforms the remaining construction methods.

\begin{table}[t]
	\caption{Results on different strategies of constructing $\mathbf{V}$, including Gram-Schmidt (\textit{G}), Cayley Transform (\textit{C}) and Birdal~\etal~\cite{birdal2018bayesian} (\textit{B}) for camera localization \ref{tab:ab_V_cam} and point cloud pose estimation \ref{tab:ab_V_pc}.}
    \begin{subtable}[h]{\linewidth}
	\begin{center}
		  \resizebox{\linewidth}{!}{
		\setlength{\tabcolsep}{5.0pt}
			\begin{tabular}{cccc}
			    \toprule
				   & UBN & MBN-MB & MBN \\
				 \cmidrule(l){2-4}
				  Threshold & G / C / B & G / C / B & G / C / B \\
				\noalign{\smallskip}
				\midrule
				\noalign{\smallskip}
				10$^\circ$ / 0.1m & 0.15 / \textbf{0.16} / 0.11 & \textbf{0.13} / \textbf{0.13} / 0.08 &  0.21 / 0.18 / \textbf{0.22}\\
				15$^\circ$ / 0.2m & 0.42 / \textbf{0.43} / 0.36 & 0.30 / \textbf{0.38} / 0.28 &  0.51 / 0.46 / \textbf{0.57}\\
				20$^\circ$ / 0.3m & \textbf{0.54} / \textbf{0.54} / 0.50 & 0.39 / \textbf{0.49} / 0.37 &  0.63 / 0.56 / \textbf{0.69}\\
				\bottomrule
			\end{tabular}
			}
		\caption{Ratio of correct poses for several thresholds, averaged over our ambiguous relocalization dataset.}
		\label{tab:ab_V_cam}
	\end{center}
	\end{subtable}
	\begin{subtable}[h]{\linewidth}
	  \centering
	  \resizebox{\linewidth}{!}{
    \begin{tabular}{ccc}
    \toprule
    UBN   & MBN-MB & MBN \\
    \midrule
    G / C / B & G / C / B & G / C / B \\
    4.654 / 2.821 / \textbf{1.477} & 3.233 / 2.458 / \textbf{1.237} & 2.867 / 1.597 / \textbf{0.973} \\
    \bottomrule
    \end{tabular}%
    }
      \caption{Average chamfer distances~($\times 10^{-2}$) on point cloud pose estimation, averaged over ModelNet10 dataset.}%
  \label{tab:ab_V_pc}%
	\end{subtable}
	\label{tab:ab_V}
\end{table}

\subsection{Variants of Multiple Hypotheses Prediction Strategies}

In our main experiments, we stick with the relaxed version of WTA, termed as RWTA.
Recently,~\cite{makansi2019overcoming} proposed EWTA, an evolving version of
WTA, to further alleviate the collapse problems of the RWTA training schemes proposed
in~\cite{rupprecht2017learning}. Updating the top $k$ hypotheses instead of
only the best one, EWTA increases the number of hypotheses that are actually
used during training, resulting in fewer wrong mode predictions that do not
match the actual distribution.
\insertimageStar{1}{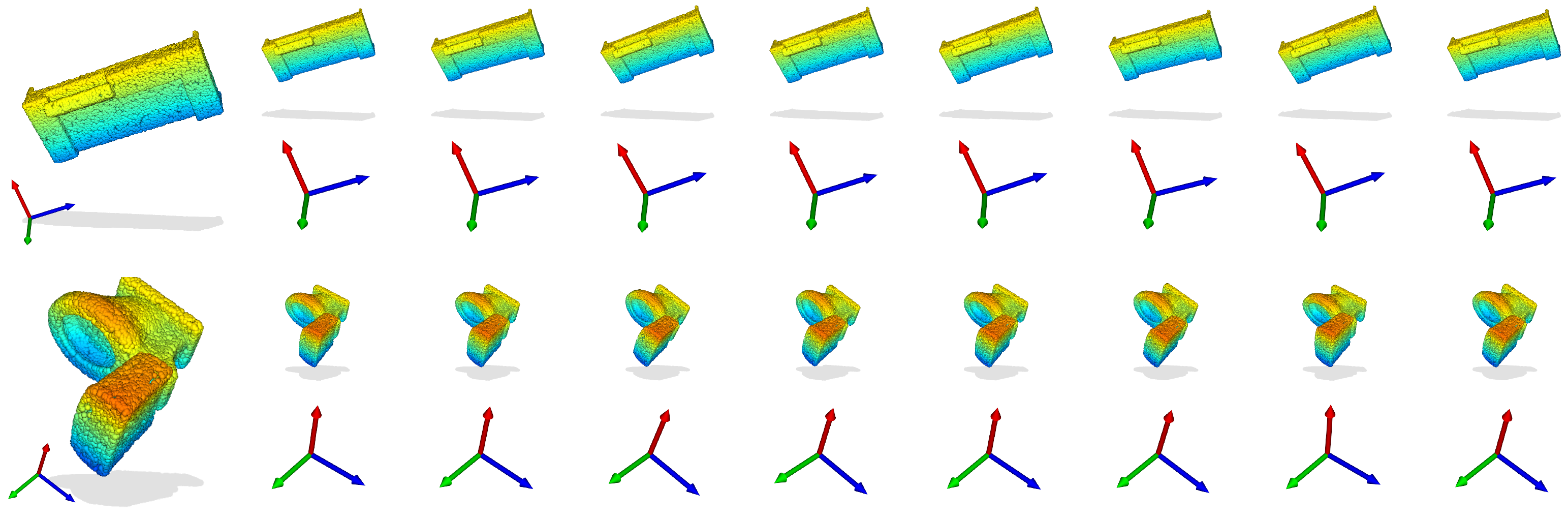}{For non-ambiguous objects, different components would generate similar predictions, where all modes correctly collapse. This can also be used to check the existence of ambiguities.}{fig:multi_pred_no_ambi}{t!}

We compared different versions of MHP training schemes for our applications, including WTA, RWTA and EWTA. The results can be
found in~\cref{tab:ab_mhp_cam} and~\cref{tab:ab_mhp_pc}.

As it is not straightforward how $k$ should be chosen in EWTA, we 1) start with $k=K$, where $K$ is the number of hypotheses and gradually decrease $k$ until $k=1$ (as proposed in \cite{makansi2019overcoming}) and 2) start with the best half hypotheses, i.e. $k=0.5 \cdot K$. We set $K=50$ in our experiments. We have found $k$ to strongly influence the accuracy of our model. In our applications, however, we have found the wrong predictions to have high uncertainty so that, if desired, they can easily be removed. Overall, RWTA results in the highest performance across all three, in both of the application scenarios. Therefore, we chose to remain with RWTA to train our models. This implicitly admits $k=1$.

\subsection{Impact of RWTA Loss}

To show how the RWTA loss helps overcome problems of mode collapse as well as facilitate the training process for MBN, we introduce a training instance following the MDN \cite{bishop1994mixture} using only Mixture Bingham Loss, dubbed as \emph{MBN-MB}. From  ~\cref{table:ambigious} and ~\cref{tab:benchmark}, we can see for both camera relocalization and point cloud pose estimation, MBN-50 demonstrates constant improved performances over MBN-MB with the same configuration. ~\cref{tab:ab_mhp} lists a complete comparison with MBN trained with only MB loss and the other versions combined with one of the WTA-based loss. We can see WTA-based losses not only improve the performance, but also increases the diversity in the multiple predictions which is illustrated by larger SEMD values. 

\begin{table}[t]
  \caption{Comparison between different MHP variants, including WTA, EWTA\cite{makansi2019overcoming} with RWTA\cite{rupprecht2017learning} for camera localization \ref{tab:ab_mhp_cam} and point cloud pose estimation \ref{tab:ab_mhp_pc}.}
  \begin{subtable}[h]{\linewidth}
    \centering
			\begin{tabular}{lccccc}
			    \toprule
			    \multirow{2}{*}{Threshold} & MB & WTA & \multirow{2}{*}{\shortstack[c]{EWTA\\(k=50)}} & \multirow{2}{*}{\shortstack[c]{EWTA\\(k=25)}} & \multirow{2}{*}{\shortstack[c]{RWTA\\(k=1, used)}} \\
				\noalign{\smallskip}\noalign{\smallskip}\noalign{\smallskip}
				\midrule
				10$^\circ$ / 0.1m & 0.08 & 0.21 & 0.20 & 0.20 & \textbf{0.22}\\
				15$^\circ$ / 0.2m & 0.28 & 0.56 & 0.51 & 0.51 & \textbf{0.57}\\
				20$^\circ$ / 0.3m & 0.37 & 0.69 & 0.62 & 0.66 & \textbf{0.69}\\
				\bottomrule
			\end{tabular}%
      \caption{Percentage of correct poses for several thresholds, averaged over all scenes of the ambiguous relocalization dataset.\vspace{2mm}}%
		\label{tab:ab_mhp_cam}%

  \end{subtable}
  \begin{subtable}[h]{\linewidth}
  \centering
    \begin{tabular}{cccccc}
    \toprule
          & MB    & WTA   & EWTA(25) & EWTA(50) & RWTA \\
    \midrule
    SEMD  & 0.425 & \textbf{0.777} & 0.639 & 0.513 & 0.729 \\
    \midrule
    CD    & 1.237 & 1.105 & 1.090 & 1.160 & \textbf{0.973} \\
    \bottomrule
    \end{tabular}%
  \caption{Average SEMD and chamfer distances $(\times 10^{-2})$ on ModelNet10 across all the classes. }%
  \label{tab:ab_mhp_pc}%
  \end{subtable}
  \label{tab:ab_mhp}
\end{table}

\begin{table}[t]
	\caption{Results using continuous 6D representation~\cite{zhou2019continuity} to model rotations instead of a Bingham distribution on the quaternion for camera localization \ref{tab:ab_6d_cam} and point cloud pose estimation \ref{tab:ab_6d_pc}}
	\begin{subtable}[h]{\linewidth}
		\centering
		\resizebox{1.0\textwidth}{!}{		
			\begin{tabular}{cccccccc}
			    \toprule
				\textbf{6D + 3D} & Threshold & Geo+L1 & Uni. & MDN & \vtop{\hbox{\strut ~~~MC- }\hbox{\strut Dropout}} & MBN-CE & MBN\\
				\noalign{\smallskip}
				\midrule
				\noalign{\smallskip}
				& 10$^\circ$ / 0.1m & 0.41 & \textbf{0.48} & 0.01 & 0.26 & 0.38 & 0.38\\
				A &15$^\circ$ / 0.2m & \textbf{0.90} & 0.89 & 0.14 & 0.83 & 0.81 & 0.79\\
				&20$^\circ$ / 0.3m & \textbf{0.96} & 0.92 & 0.23 & 0.91 & 0.84 & 0.81\\
				\midrule
				&10$^\circ$ / 0.1m & 0.03 & 0.03 & 0.02 & 0.02 & 0.06 & \textbf{0.07}\\
				B &15$^\circ$ / 0.2m & 0.16 & 0.16 & 0.11 & 0.13 & 0.29 & \textbf{0.33}\\
				&20$^\circ$ / 0.3m & 0.22 & 0.23 & 0.14 & 0.21 & 0.38 & \textbf{0.42}\\
				\midrule
				&10$^\circ$ / 0.1m & 0.17 & 0.19 & 0.12 & 0.12 & 0.18 & \textbf{0.20}\\
				C &15$^\circ$ / 0.2m & 0.46 & \textbf{0.51} & 0.36 & 0.36 & 0.44 & 0.49\\
				&20$^\circ$ / 0.3m & 0.62 & \textbf{0.67} & 0.47 & 0.56 & 0.56 & 0.61\\
				\midrule
				&10$^\circ$ / 0.1m & 0.07 & 0.01 & 0.01 & 0.04 & \textbf{0.08} & \textbf{0.08}\\
				D &15$^\circ$ / 0.2m & 0.30 & 0.06 & 0.09 & 0.18 & 0.35 & \textbf{0.40}\\
				&20$^\circ$ / 0.3m & 0.48 & 0.13 & 0.14 & 0.36 & 0.55 & \textbf{0.60}\\
				\midrule
				&10$^\circ$ / 0.1m & 0.34 & 0.24 & 0.30 & 0.21 & 0.34 & \textbf{0.42}\\
				E &15$^\circ$ / 0.2m & 0.74 & 0.63 & 0.65 & 0.65 & 0.76 & \textbf{0.77}\\
				&20$^\circ$ / 0.3m & 0.84 & 0.79 & 0.76 & 0.82 & 0.88 & \textbf{0.90}\\
				\midrule
				&10$^\circ$ / 0.1m & 0.20 & 0.19 & 0.09 & 0.13 & 0.21 & \textbf{0.23}\\
				Average &15$^\circ$ / 0.2m & 0.51 & 0.45 & 0.27 & 0.43 & 0.53 & \textbf{0.56}\\
				&20$^\circ$ / 0.3m & 0.63 & 0.55 & 0.35 & 0.57 & 0.64 & \textbf{0.67}\\
				\bottomrule
			\end{tabular}
		}
		\caption{Ratio of correct camera poses on our ambiguous relocalization dataset. Camera poses are modelled with six dimensions for rotation and three for translations resulting in a nine dimensional representation for a camera pose. A: Blue Chairs, B: Meeting Table, C: Staircase, D: Staircase Extended and E: Seminar Room.}
		\label{tab:ab_6d_cam}
	\end{subtable}
	\begin{subtable}[h]{\linewidth}
	\centering
	\resizebox{1.0\textwidth}{!}{

	    \begin{tabular}{ccccccc}
    \toprule
        \textbf{6D}  & Geo   & Uni   & MDN   & MC-Dropout & MBN-CE & MBN \\
    \midrule
    Bathtub & 6.246 & 7.228 & 0.513 & 4.048 & 0.445 & \textbf{0.427} \\
    Bed   & 1.533 & 1.730 & 0.517 & 1.407 & \textbf{0.277} & 0.346 \\
    Chair & 0.559 & 0.702 & \textbf{0.515} & 0.774 & 0.543 & 0.537 \\
    Desk  & 4.404 & 4.650 & 3.144 & 4.050 & \textbf{3.023} & 3.193 \\
    Dresser & 4.792 & 3.799 & 2.216 & 2.590 & \textbf{2.121} & 2.333 \\
    Monitor & 1.694 & 2.124 & \textbf{1.076} & 2.136 & 1.178 & 1.161 \\
    Night Stand & 3.756 & 3.656 & 1.559 & 2.502 & 1.445 & \textbf{1.434} \\
    Sofa  & 0.313 & 0.374 & 0.319 & 0.503 & 0.324 & \textbf{0.312} \\
    Table & 9.913 & 8.685 & \textbf{0.473} & 3.161 & 0.654 & 0.621 \\
    Toilet & 0.427 & 0.873 & 1.065 & 0.750 & \textbf{0.266} & 0.423 \\
    \midrule
    Average & 3.364 & 3.382 & 1.140 & 2.192 & \textbf{1.028} & 1.079 \\
    \bottomrule
    \end{tabular}%
    }
	\caption{Point cloud pose estimation results on ModelNet10 across all the classes using the 6D representation~\cite{zhou2019continuity} to model rotations.}
	\label{tab:ab_6d_pc}%
	\end{subtable}
	\label{tab:ab_6d}
\end{table}

\subsection{Variants of Rotation Parameterization}
The best choice of rotation parameterization for training deep learning models is an open question. PoseNet~\cite{kendall2015posenet} proposed to use quaternions due to the ease of normalization. The ambiguities can be resolved by mapping the predictions to one hemisphere. MapNet~\cite{brahmbhatt2018geometry} further showed improvements in using the axis angle representation. Recently it has been shown that any representation with four or less degrees of freedom suffers from discontinuities in mapping to $SO(3)$. This might harm the performance of deep learning models. Instead,~\cite{zhou2019continuity} proposed a continuous 6D or 5D representation. We ablate in this context by mapping all predictions to the proposed 6D representation and model them using a Gaussian mixture models (GMM), similar to a MDN. In  ~\cref{tab:ab_6d_cam}, 'Geo + L1' refers to a direct regression using the geodesic loss proposed in \cite{zhou2019continuity} and an $l_1$ loss on the translation. ~\cref{tab:ab_6d_pc} lists results for point cloud pose estimation. In both cases, we can see there is a clear improvement in the results when the model is lifted from a single prediction to multiple predictions, which further validates ambiguities could be well handled with multimodality. Also, in both MBN variants, when RWTA loss is incorporated, the performence could be further boosted from pure MDN, which qualifies our proposed training schemes as well.

\section{Conclusion}
\label{sec:conclude}

We have proposed an elegant solution in this paper to enable end-to-end modeling of pose distributions for 3D rotation estimation via deep networks. We cover both unimodal (UBN) and multimodal (MBN) cases that are able to infer single as well as mixture distributions. We illustrate the feasibility to train neural networks to regress parameters of a Bingham distribution for obtaining pose predictions as well as uncertainty information. A MDN-like Multimodal Bingham network is designed targeting ambiguity issues which cannot be handled by Unimodal Bingham Network. Novel training schemes which resort to WTA strategy to facilitate the success of training a Bingham Mixture model are evaluated, avoiding mode collapse, and as a result providing multiple plausible pose predictions as well as associated uncertainty values in each hypothesis. We exhaustively evaluated our methods on two fundamental pose-related vision tasks, namely point cloud pose estimation and camera localization. For the latter we extend our framework to additionally model the translation of a camera's pose using Gaussian Mixture Models. We demonstrated our model's superiority over the state-of-the-art on both tasks, obtaining consistently better mode predictions. We believe those solutions can be easily incorporated into other neural network-based pose estimation applications to improve their performances without heavy modifications.    

\begin{acknowledgements}
This project is supported by Bavaria California Technology Center (BaCaTeC), Stanford-Ford Alliance, NSF grant IIS-1763268, Vannevar Bush Faculty Fellowship, Samsung GRO program, the Stanford SAIL Toyota Research, and the PRIME programme of the German Academic Exchange Service (DAAD) with funds from the German Federal Ministry of Education and Research (BMBF).
\end{acknowledgements} \vspace{-10mm}

\bibliographystyle{spmpsci}      % mathematics and physical sciences
\bibliography{egbib}

\end{document}